\documentclass[final,3p,times]{elsarticle}

\usepackage{subfigure}
\usepackage{amssymb}
\usepackage{amsmath}
\usepackage{multirow, booktabs}

\usepackage{ulem}

\usepackage{color, soul}
\usepackage{hyperref}
\hypersetup{colorlinks=true,
            linkcolor=blue,
            anchorcolor=blue,
            citecolor=blue}

\journal{Pattern Recognition}

\begin{document}

\begin{frontmatter}



\title{Robust Table Structure Recognition with Dynamic Queries Enhanced Detection Transformer}

\author[ustc,msra]{Jiawei Wang\corref{leis}\fnref{myfootnote}}
\ead{wangjiawei@mail.ustc.edu.cn}

\author[msra]{Weihong Lin}
\ead{lwher1996@outlook.com}

\author[msra]{Chixiang Ma}
\ead{ma1996@mail.ustc.edu.cn}

\author[sjtu]{Mingze Li\fnref{myfootnote}}
\ead{yagami@sjtu.edu.cn}

\author[ucas]{Zheng Sun\fnref{myfootnote}}
\ead{sunzheng2019@gmail.com}

\author[msra]{Lei Sun\corref{leis}}
\ead{kuangtongustc@gmail.com}

\author[msra]{Qiang Huo}
\ead{qianghuo@microsoft.com}

\affiliation[ustc]{organization={University of Science and Technology of China},
            city={Hefei},
            postcode={230026}, 
            country={China}}

\affiliation[msra]{organization={Microsoft Research Asia},
            city={Beijing},
            postcode={100080}, 
            country={China}}

\affiliation[sjtu]{organization={Shanghai Jiao Tong University},
            city={Shanghai},
            postcode={200240}, 
            country={China}}

\affiliation[ucas]{organization={University of Chinese Academy of Sciences},
            city={Beijing},
            postcode={100049}, 
            country={China}}


\fntext[myfootnote]{Work done when Jiawei Wang, Mingze Li and Zheng Sun were interns at MMI Group, Microsoft Research Asia, Beijing, China.}
\cortext[leis]{Corresponding author.}

\begin{abstract}
We present a new table structure recognition (TSR) approach, called TSRFormer, to robustly recognize the structures of complex tables with geometrical distortions from various table images. Unlike previous methods, we formulate table separation line prediction as a line regression problem instead of an image segmentation problem and propose a new two-stage dynamic queries enhanced DETR based separation line regression approach, named DQ-DETR, to predict separation lines from table images directly. Compared to Vallina DETR, we propose three improvements in DQ-DETR to make the two-stage DETR framework work efficiently and effectively for the separation line prediction task: 1) A new query design, named Dynamic Query, to decouple single line query into separable point queries which could intuitively improve the localization accuracy for regression tasks; 2) A dynamic queries based progressive line regression approach to progressively regressing points on the line which further enhances localization accuracy for distorted tables; 3) A prior-enhanced matching strategy to solve the slow convergence issue of DETR. After separation line prediction, a simple relation network based cell merging module is used to recover spanning cells. With these new techniques, our TSRFormer achieves state-of-the-art performance on several benchmark datasets, including SciTSR, PubTabNet, WTW, FinTabNet, and cTDaR TrackB2-Modern. Furthermore, we have validated the robustness and high localization accuracy of our approach to tables with complex structures, borderless cells, large blank spaces, empty or spanning cells as well as distorted or even curved shapes on a more challenging real-world in-house dataset.
\end{abstract}




\end{frontmatter}


\section{Introduction}
\label{sec:intro}
Tables offer a means to efficiently represent and communicate structured data in many scenarios like scientific publications, financial statements, invoices, web pages, etc. Due to the trend of digital transformation, automatic table structure recognition (TSR) has become an important research topic in document understanding and attracted the attention of many researchers. TSR aims to reconstruct the cellular structures of tables from table images by extracting the coordinates and row/column spanning information of cell boxes. This task is very challenging since tables may have complex structures, diverse styles, and contents, and become geometrically distorted or even curved during the image-capturing process.

\begin{figure}[htbp]
    \centering
    \subfigure[RobusTabNet\cite{ma2022robust}]{
        \includegraphics[width=5cm]{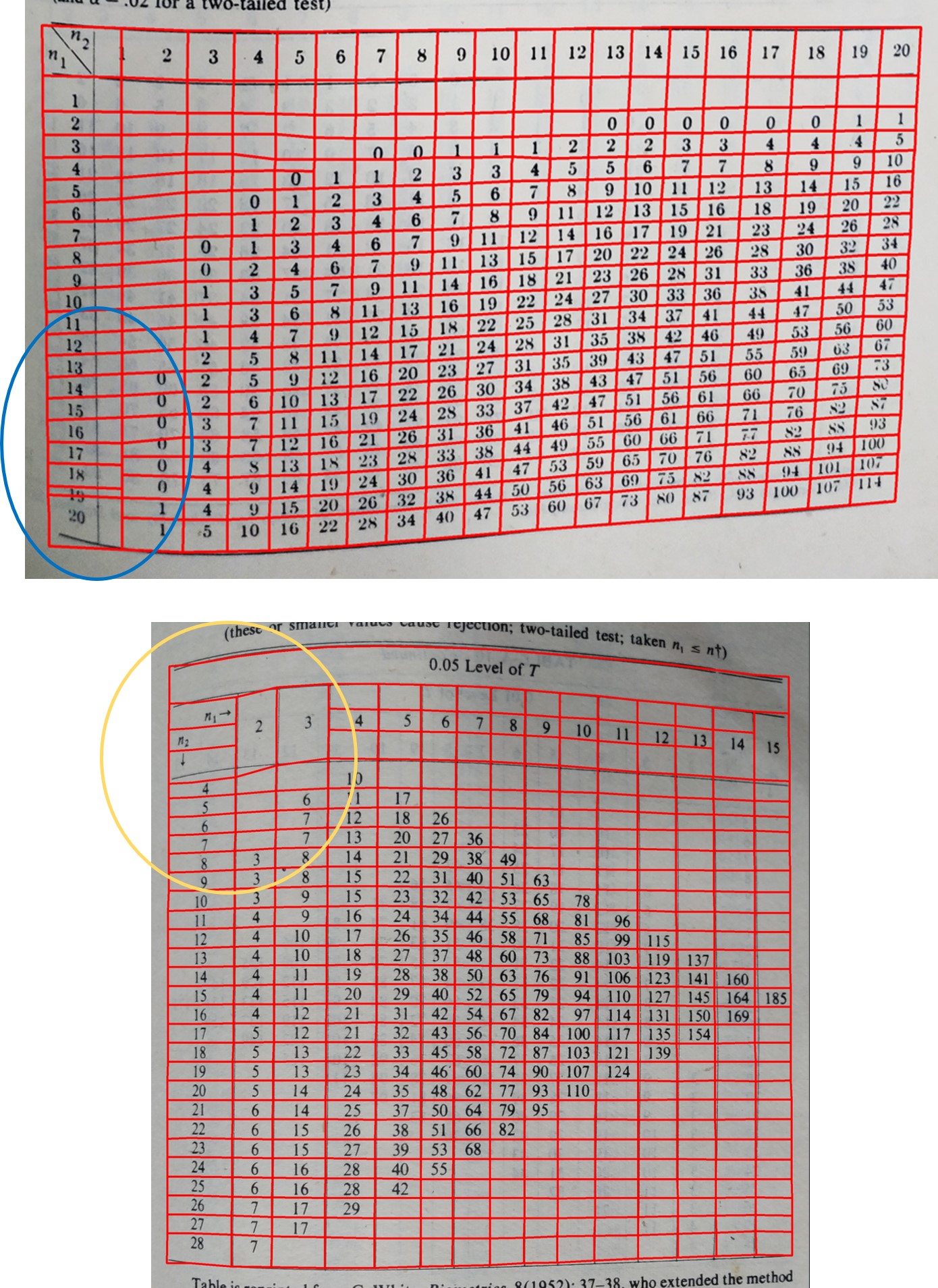}
    }
    \subfigure[TSRFormer with SepRETR\cite{lin2022tsrformer}]{
        \includegraphics[width=5cm]{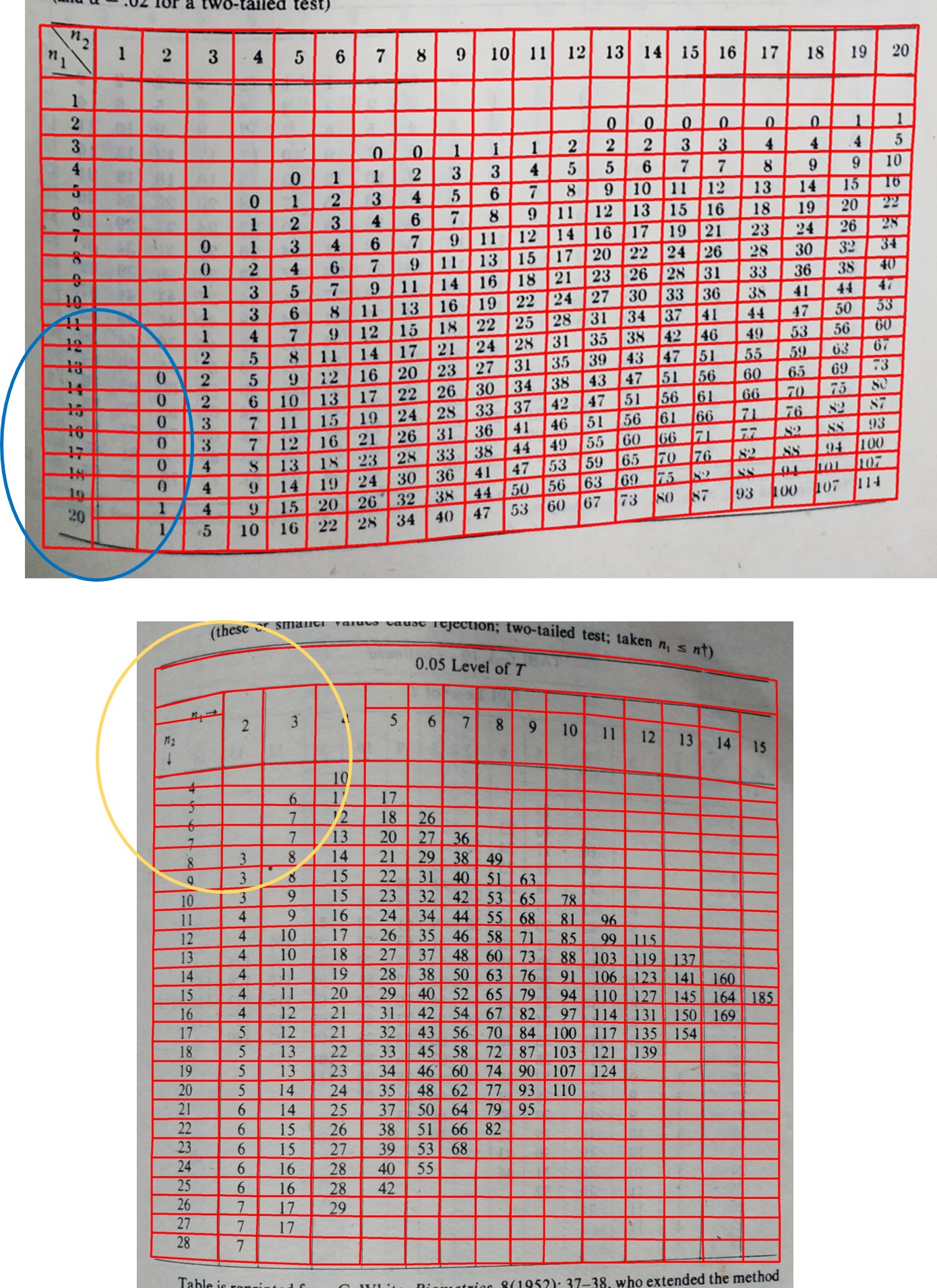}
    }
    \subfigure[TSRFormer with DQ-DETR]{
        \includegraphics[width=5cm]{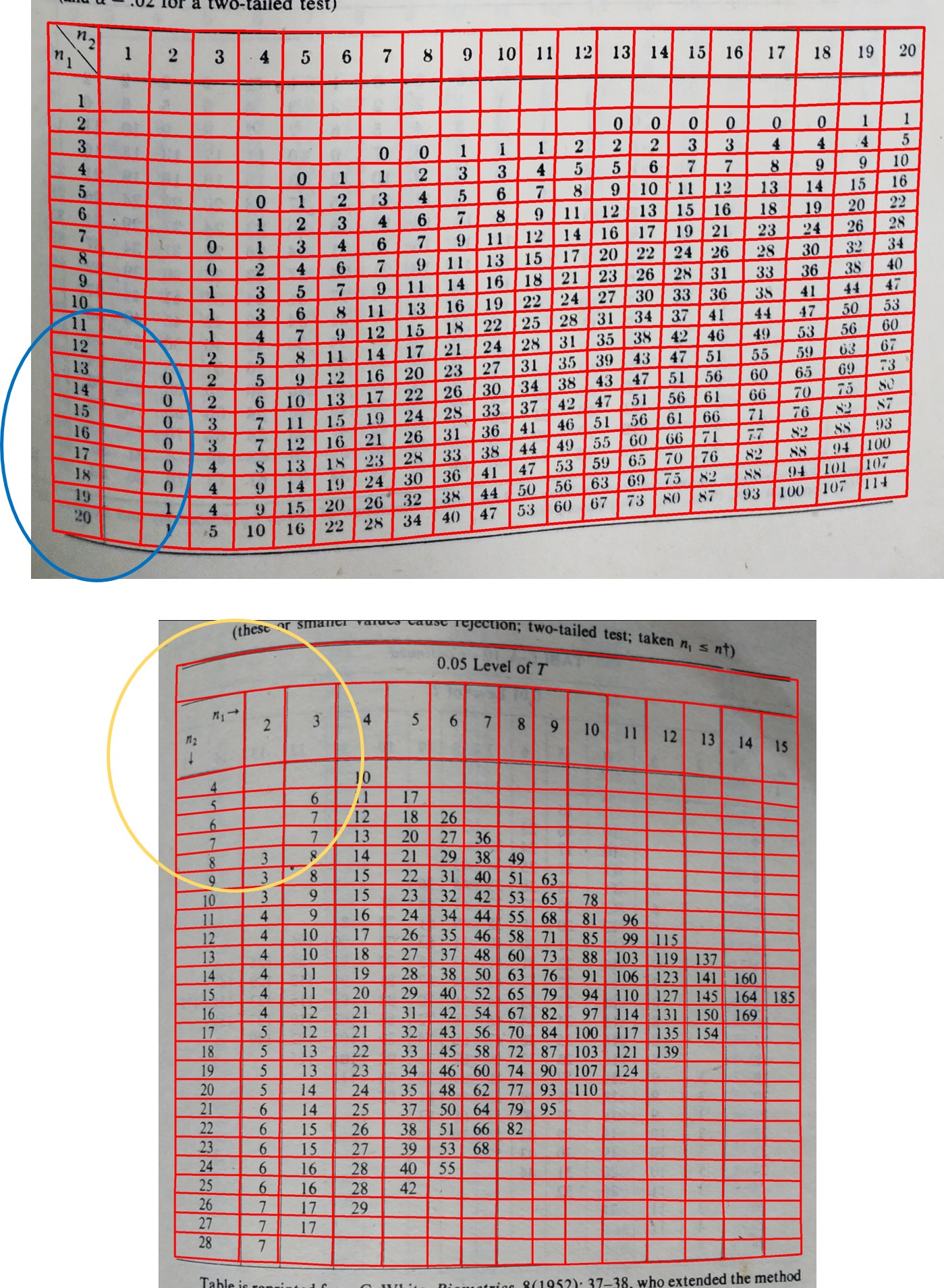}
    }
    \centering
    \caption{Comparison of the results of RobusTabNet, SepRETR based TSRFormer, and DQ-DETR based TSRFormer on two challenging distorted tables. }
    \label{fig-comp}
\end{figure}

In recent years, deep learning based TSR methods, e.g., \cite{deepdesrt2017,paliwal2019tablenet,deeptabstr2019,rethinking2019,rethinkinggnn2019,SPLERGE,TabStruct2020,GTE2021,xue2021tgrnet,long2021parsing,FLAG2021,Qiao2021LGPMACT}, have made impressive progress towards recognizing the structures of tables detected in scanned documents or PDF files. However, how to robustly recognize the structures of geometrically distorted or even curved tables, which appear often in camera-captured images, is still an under-researched problem. Only several very recent works made some attempts to overcome this challenge. For instance, Cycle-CenterNet \cite{long2021parsing} proposed an effective approach to parsing the structures of distorted bordered tables in wild complex scenes and achieved promising results on their WTW \cite{long2021parsing} dataset, but this method cannot perform well for borderless tables. NCGM \cite{liu2022neural} found that previous graph based TSR methods cannot process distorted tables reliably and proposed a new method called Neural Collaborative Graph Machines (NCGM) to leverage inter-intra modality collaboration to enhance the embeddings of text segments in each table, based on which the row/column/cell grouping relationships between text segments in distorted tables can be predicted more robustly.  Nevertheless, this method relies on using an OCR engine to extract text segment bounding boxes and contents from table images first, so it is not robust to tables with a number of undetected text segments or empty cells. Unlike NCGM, the latest split-and-merge based TSR method, namely RobusTabNet\cite{ma2022robust}, does not depend on OCR results.    This approach proposed to leverage spatial CNN modules to enhance the
feature representation of each pixel on convolutional feature
maps by propagating contextual information across the whole feature map in horizontal or vertical directions, which can significantly improve the robustness of semantic segmentation based separation line prediction models to distorted (even curved) tables. Although having achieved promising results on a real-world dataset containing both distorted and undistorted tables, RobusTabNet still struggles with some challenging cases, e.g., distorted tables with many empty cells, which are shown in Fig.~\ref{fig-comp}. This is because, even if enhanced by spatial CNN modules, the separation line segmentation model still cannot produce high-quality segmentation masks for these challenging cases.

\begin{figure}[htbp]
    \centering
    \subfigure[Direct line regression \cite{lin2022tsrformer}]{
        \includegraphics[width=7cm]{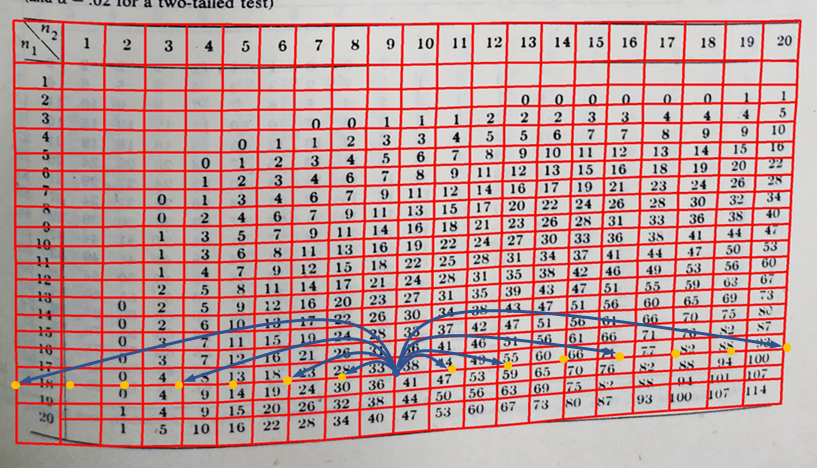}
    }
    \subfigure[Progressive line regression]{
        \includegraphics[width=7cm]{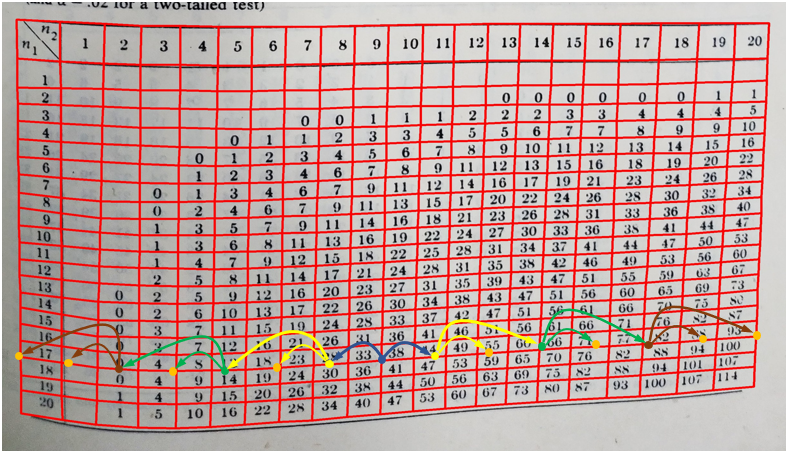}
    }
    \centering
    \caption{Comparison of direct line regression in SepRETR \cite{lin2022tsrformer} and progressive line regression in DQ-DETR. }
    \label{fig-regression}
\end{figure}

In this paper, we propose a new split-and-merge based TSR approach, called TSRFormer, to recognize the structures of various tables from table images robustly. TSRFormer contains two effective components: 1) A two-stage DETR based separator regression module to directly predict linear and curvilinear row/column separation lines from input table images; 2) A relation network based cell merging module to recover spanning cells by merging adjacent cells generated by intersecting row and column separators. Unlike previous split-and-merge based approaches like RobusTabNet, we formulate table separation line prediction as a line regression problem instead of an image segmentation problem. To this end, we have introduced a new two-stage DETR \cite{deformdetr2021} based separation line prediction approach in our conference paper \cite{lin2022tsrformer}, dubbed Separator REgression TRansformer (SepRETR), to detect separation lines from table images directly. Specifically, SepRETR tries to detect a reference point for each separation line first. Then, a DETR decoder takes the embeddings of these reference points as input queries and leverages cross-attention and self-attention operators in each decoder layer to enhance the embedding of each query. Finally, each enhanced query embedding is input into a classifier to predict whether this query is a false alarm or not, and each remaining query embedding is further fed into a regressor to regress the positions of other points on its corresponding separation line directly. Compared with RobusTabNet, SepRETR has two advantages: 1) SepRETR doesn't rely on using heuristic post-processing algorithms to convert separation line segmentation masks into separation lines; 2) SepRETR can predict separation lines more robustly especially when dealing with distorted tables. However, we find that SepRETR cannot regress the positions of points distant from reference points as precisely as that of points near reference points in some challenging cases shown in Fig.~\ref{fig-comp}. The reason is that the cross-attention operators in the decoder tend to assign lower attention scores to pixels distant from reference points so that the enhanced embedding of each reference point doesn't contain enough information to predict the positions of distant points in these challenging cases precisely. Based on this observation, we propose a new progressive separation line regression algorithm to achieve higher line regression accuracy. As illustrated in Fig.~\ref{fig-regression}, instead of regressing the positions of other points on each separation line all at once, we propose a new DETR based separation line prediction model, dubbed Dynamic Queries enhanced DETR (DQ-DETR), to regress the positions of points on each separation line progressively. Unlike previous DETR models, the number of queries in each DQ-DETR decoder layer is not fixed. 
Specifically, given an initial set of reference points detected from the input table image, the first decoder layer takes them as input queries and generates an enhanced embedding for each query, followed by refining the positions of their corresponding points using a regressor. Then, one additional point is appended on the left and right sides of its corresponding reference point respectively, serving as additional queries for the second decoder layer to detect new points. At the subsequent layers of the decoder, two more points are appended on both sides of the detected separation lines based on the positions of the existing points after each layer. These new points are also taken as new queries for the next decoder layer to refine their positions.
This regression algorithm is done iteratively to obtain all the points on each separation line. In each decoder layer, the regressor only needs to refine the positions of new points near the reference points, so the regression accuracy can be improved significantly. As illustrated in Fig.~\ref{fig-comp}, DQ-DETR is much more robust to distorted tables than SepRETR. Moreover, we propose two effective techniques to improve the efficiency of DQ-DETR in both training and inference: 1) A prior-enhanced matching strategy to accelerate the convergence speed of DQ-DETR; 2) Leveraging the factorized self-attention \cite{dong2021visual} and deformable attention \cite{deformdetr2021} modules to replace the original self-attention and cross-attention modules in DETR decoder layers to significantly reduce the computation cost of DQ-DETR in inference. With the help of DQ-DETR, our TSRFormer has achieved
state-of-the-art performance on several public TSR benchmarks,
including SciTSR \cite{chi2019complicated}, PubTabNet \cite{zhong2020image}, WTW \cite{long2021parsing}, FinTabNet \cite{zheng2020global}, and cTDaR TrackB2-Modern \cite{gao2019icdar}. Furthermore,
we have demonstrated the robustness of our approach to tables
with complex structures, borderless cells, large blank spaces, empty or spanning cells as well as distorted or even curved shapes on a
more challenging real-world in-house dataset.

The main contributions of this paper are as follows:
\begin{itemize}
\item To the best of our knowledge, we are the first to formulate separation line prediction as a line regression problem. To demonstrate the effectiveness of this new formulation, we propose a new DETR based separation line prediction model, dubbed DQ-DETR, to predict the positions of points on separation lines from both distorted and undistorted table images with high localization accuracy.

\item We introduce the concept of dynamic queries into the DETR framework, thanks to which our new progressive separation line regression algorithm can be implemented in the DETR framework efficiently. Moreover, we propose a new prior-enhanced matching strategy to accelerate the convergence speed of DQ-DETR in training and leverage the factorized self-attention \cite{dong2021visual} and deformable attention \cite{deformdetr2021} modules to significantly reduce the computation cost of DQ-DETR in inference further. 

\item With the help of DQ-DETR, our TSRFormer has achieved state-of-the-art performance on several public TSR benchmarks, including SciTSR \cite{chi2019complicated}, PubTabNet \cite{zhong2020image}, WTW \cite{long2021parsing}, FinTabNet \cite{zheng2020global}, and cTDaR TrackB2-Modern \cite{gao2019icdar}. 

\end{itemize}

Although a preliminary study of TSRFormer has been presented in our conference paper \cite{lin2022tsrformer}, this paper extends it significantly in  the following aspects: (1) A new DETR based separation line regression model, named DQ-DETR, is proposed to predict the positions of points on each separation line progressively, which leads to higher separation line prediction accuracy; (2) More ablation studies are conducted to demonstrate the effectiveness of DQ-DETR; (3) Experimental results on a new public benchmark dataset, namely FinTabNet \cite{zheng2020global}, are presented to compare our approach with other approaches more comprehensively.

\section{Related work}
\subsection{Table structure recognition}
Early TSR methods were mainly based on handcrafted features and heuristic rules (e.g., \cite{laurentini1992identifying,itonori1993table,trecs1998,shigarov2016configurable,rastan2019texus}), so they could only deal with simple table structures or specific data formats, such as PDF files. Later, some statistical machine learning based methods (e.g, \cite{ng1999learning,wang2004table}) were proposed to reduce the dependence on heuristic rules. However, these methods still made strong assumptions about table layouts and relied on handcrafted features, which limited their generalization ability. In recent years, many deep learning based approaches have emerged and outperformed these traditional methods significantly in terms of both accuracy and capability. These approaches can be roughly divided into three categories: row/column extraction based methods, image-to-markup generation based methods and bottom-up methods.

\textbf{Row/column extraction based methods.} These approaches leverage object detection or semantic segmentation methods to detect entire rows and columns first, then intersect them to form a grid of cells. DeepDeSRT \cite{deepdesrt2017} first applied an FCN-based semantic segmentation method \cite{fcn2015} to table structure extraction. TableNet \cite{paliwal2019tablenet} proposed an end-to-end FCN-based model to simultaneously detect tables and recognize table structures. However, these vanilla FCN-based TSR methods are not robust to tables containing large blank spaces due to limited receptive fields. To alleviate this problem, methods like \cite{rethinking2019,SPLERGE,khan2019table} tried different context enhancement techniques, e.g., pooling features along rows and columns of pixels on some intermediate feature maps of FCN models or using sequential models like bi-directional gated recurrent unit networks (GRU), to improve row/column segmentation accuracy. Another group of approaches \cite{deeptabstr2019,hashmi2021guided, smock2022pubtables} treated TSR as an object detection problem and used some object detection methods to directly detect the bounding boxes of rows and columns. Among these methods, SPLERGE \cite{SPLERGE} was the first to deal with spanning cells, which proposed to add a simple cell merging module after a row/column extraction module to recover spanning cells by merging adjacent cells. Later, several works were proposed to further improve the cell merging module. TGRNet \cite{xue2021tgrnet} designed a network to jointly predict the spatial locations and spanning information of table cells. SEM \cite{zhang2022split} fused the features of each cell from both vision and text modalities. Raja et al. \cite{raja2022visual} improved this "split-and-merge" paradigm by targeting row, column, and cell detection as object detection tasks and forming rectilinear associations through a graph-based formulation for generating row/column spanning information. Different from this two-stage paradigm, Zou et al. \cite{zou2020deep} proposed a one-stage approach to predicting the real row and column separators to handle spanning cells. Although these methods have achieved impressive performance on some previous benchmarks, e.g., \cite{gobel2013icdar,chi2019complicated,zhong2020image}, they are not able to handle distorted or curved tables because they rely on an assumption that tables are axis-aligned. For tables with rotation and linear perspective transformation, Zeng et al. \cite{guo2022trust} proposed an end-to-end transformer-based method to predict the start points and rotation angles of separation lines and merge basic grids. However, this method cannot deal with curved tables because it assumes that the separation lines of tables are straight. To make split-and-merge based TSR methods robust to distorted tables, RobusTabNet \cite{ma2022robust} proposed to incorporate spatial CNN modules into semantic segmentation based separation line prediction models to enhance the representation ability of their convolutional feature maps by propagating contextual information across the whole feature map in horizontal or vertical directions. Although having achieved promising results on a real-world dataset containing both distorted and undistorted tables, RobusTabNet still struggles with some challenging cases like distorted tables with many empty cells.


\textbf{Image-to-markup generation based methods.} This type of method treats TSR as a problem of generating markup from images and adopts existing image-to-markup models to directly convert each source table image into a target presentational markup that fully describes its structure and cell contents. Deng et al. \cite{deng2019challenges} constructed a new dataset TABLE2LATEX-450K and proposed to make use of an attentional encoder-decoder model to convert tables into LaTeX source codes. Li et al. \cite{li2020tablebank} defined a set of HTML tags to describe table structures only and presented a new table benchmark dataset known as TableBank. Zhong et al. \cite{zhong2020image} introduced another large-scale table benchmark dataset PubTabNet, which contains 568k table images with corresponding structured HTML representations, and introduced an attention-based encoder-dual-decoder architecture to recognize table structures and cell contents simultaneously. One common limitation of these methods is that they cannot provide the bounding box of each table cell in the original image. To solve this problem, some later work \cite{he2021pingan,nassar2022tableformer} designed models with different decoder branches to predict not only a sequence of tags representing the structure of each table but also the bounding boxes of table cells. All these methods rely on a large amount of data to train their models to achieve high performance. As camera-captured table images in existing datasets are scarce, the effectiveness of these methods for recognizing tables in camera-captured images has not been verified yet. To alleviate this data insufficiency issue, Chen et al. \cite{chen2022complex} proposed a new table structure representation, called Identity Matrix, and a new data augmentation method, named TabSplitter, to enhance the diversity of the training data for their encoder-decoder based table structure recognition model. Although this method has achieved promising results for complex tables in the wild, it is still difficult to deal with very large and complex tables due to the limit of the maximum length of the output sequence.

\textbf{Bottom-up methods.} Bottom-up methods can be further categorized into two groups. The first group \cite{rethinkinggnn2019,chi2019complicated,li2021gfte,xue2019res2tim,li2022table,FLAG2021} treats primitive regions like words or cell contents as nodes in a graph and uses graph neural networks to predict whether each sampled node pair is in a same cell, row or column. NCGM \cite{liu2022neural} found that these previous graph based TSR methods cannot process distorted tables reliably and proposed a new method called Neural Collaborative Graph Machines (NCGM) to leverage inter-intra modality collaboration to enhance the embeddings of text segments in each table, based on which the row/column/cell grouping relationships between text segments in distorted tables can be predicted more robustly. However, this method still relies on using an OCR engine to extract text segment bounding boxes and contents from table images, so it is not robust to tables with a number of undetected text segments or empty cells. To bypass this problem, the second group of methods \cite{GTE2021,prasad2020cascadetabnet,TabStruct2020,li2021adaptive,Qiao2021LGPMACT} detects the bounding boxes of table cells directly and uses different methods to group them into rows and columns. After cell detection, methods like \cite{GTE2021,li2021adaptive,Qiao2021LGPMACT} used heuristic rules to cluster detected cells into rows and columns. CascadeTabNet \cite{prasad2020cascadetabnet} recovered cell relations based on some rules for borderless tables and intersected detected separation lines to extract the grid of bordered tables. TabStruct-Net \cite{TabStruct2020} proposed an end-to-end network to detect cells and predict cell relations jointly. However, these approaches fail to handle tables containing a large number of empty cells or distorted/curved tables. Cycle-CenterNet \cite{long2021parsing} proposed to detect the vertices and center points of cells first and then group the cells into tabular objects by learning the common vertices. This method can parse the structures of distorted bordered tables in wild complex scenes effectively, but it cannot perform well for borderless tables. 

\subsection{DETR and its variants}
DETR \cite{detr2020} is a novel Transformer-based \cite{transformer2017} object detection algorithm, which introduced the concept of object query and set prediction loss to object detection. These novel attributes make DETR get rid of many manually designed components in previous CNN-based object detectors like anchor design and non-maximum suppression (NMS). However, DETR has its own issues: 1) Slow training convergence; 2) Unclear physical meaning of object queries; 3) Hard to leverage high-resolution feature maps due to high computational complexity. Deformable DETR \cite{deformdetr2021} proposed several effective techniques to address these issues: 1) Formulating queries as 2D anchor points; 2) Designing a deformable attention module that only attends to certain sampling points around a reference point to efficiently leverage multi-scale feature maps; 3) Proposing a two-stage DETR framework and an iterative bounding box refinement algorithm to further improve accuracy. Inspired by the concept of reference point in Deformable DETR, some follow-up works attempted to address the slow convergence issue by giving spatial priors to the object query. For instance, Conditional DETR \cite{meng2021conditional} divided the cross-attention weights into two parts, i.e., content attention weights and spatial attention weights, and proposed a conditional spatial query to make each cross-attention head in each decoder layer focus on a different part of an object. Anchor DETR \cite{wang2022anchor} generated object queries from 2D anchor points directly. DAB-DETR \cite{liu2022dab} proposed to use 4D anchor box coordinates to represent queries and dynamically update boxes in each decoder layer. SMCA \cite{gao2021fast} first predicted a reference 4D box for each query and then directly generated its related spatial cross-attention weights with a Gaussian prior in the transformer decoder. Inspired by two-stage Deformable DETR, Efficient DETR \cite{efficientdetr} took top-K scored proposals output from the first dense prediction stage and their encoder features as the reference boxes and object queries, respectively. Different from the above works, TSP \cite{sun2021rethinking} discarded the whole DETR decoder and proposed an encoder-only DETR. DN-DETR \cite{li2022dn} pointed out that the bipartite matching algorithm used in Hungarian loss is another reason for slow convergence and proposed a denoising based training method to speed up DETR convergence. DINO \cite{zhang2022dino} improved the performance and efficiency of DN-DETR further by using a contrastive way for denoising training, a mixed query selection method for anchor initialization, and a look-forward-twice scheme for box prediction.

\section{Methodology}

\begin{figure*}
    \centering
    \includegraphics[width=1.0\textwidth]{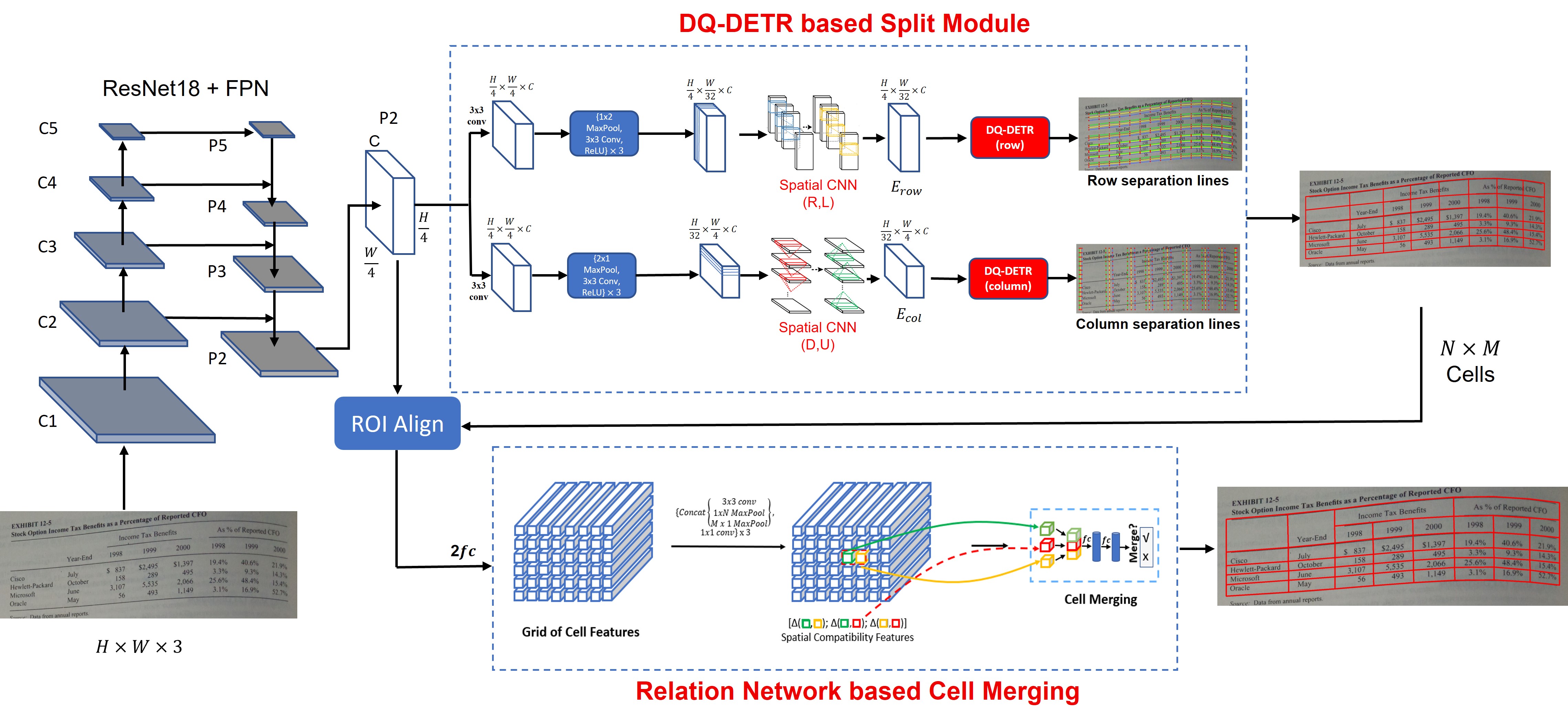}
    \caption{An overview of the proposed TSRFormer.}
    \label{fig-1}
\end{figure*}

\subsection{Overview}
As depicted in Fig.~\ref{fig-1}, the improved TSRFormer model contains two key components: 1) A DQ-DETR based split module to predict all row and column separation lines from each input table image; 2) A relation network based cell merging module \cite{ma2022robust} to recover spanning cells. These two modules are attached to a shared convolutional feature map $P_2$ generated by a ResNet18-FPN backbone \cite{resnet,fpn2017}. Details of these two components will be 
described in Section \ref{subsec: DQ-DETR} and Section \ref{subsec:relation-network}, respectively.

\subsection{DQ-DETR based split module}
\label{subsec: DQ-DETR}
In the split module, two parallel branches are attached to the shared feature map $P_2$ to predict row and column separators, respectively. Each branch comprises two modules: (1) A spatial CNN based feature enhancement module \cite{ma2022robust} to generate a context-enhanced feature map; (2) A DQ-DETR decoder to predict the positions of all separation lines. In subsequent sections, we will take the row separation line prediction branch as an example to introduce the details of these two modules.

\subsubsection{Spatial CNN based feature enhancement}
Following RobusTabNet \cite{ma2022robust}, a spatial CNN based feature enhancement module is used to enhance the feature representation of each pixel on $P_2$ first. As shown in Fig.~\ref{fig-1}, we add a $3\times3$ convolutional layer and three repeated down-sampling blocks, each composed of a sequence of a $1\times2$ max-pooling layer, a $3\times3$ convolutional layer, and a ReLU activation function, after $P_2$ sequentially to generate a down-sampled feature map $P_2'\in R^{\frac{H}{4}\times\frac{W}{32}\times C}$ first. Then, two cascaded spatial CNN (SCNN) \cite{spatialcnn} modules are attached to $P_2'$ to enhance its feature representation ability further by propagating contextual information across the whole feature map in rightward and leftward directions. Take the rightward direction as an example, the SCNN module splits $P_2'$ into $\frac{W}{32}$ slices along the width direction and propagates the information from the leftmost slice to the rightmost slice sequentially with convolution operators. Specifically, each slice is convolved by a convolutional layer with the kernel size of 9 × 1 (9 and 1 represent kernel height and width respectively) and the output feature map is fused with its right slice by element-wise addition. The output context-enhanced feature map $E_{row}$ is taken as the input of the following DQ-DETR decoder.


\begin{figure}[ht]
  \centering
  \includegraphics[width=0.9\linewidth]{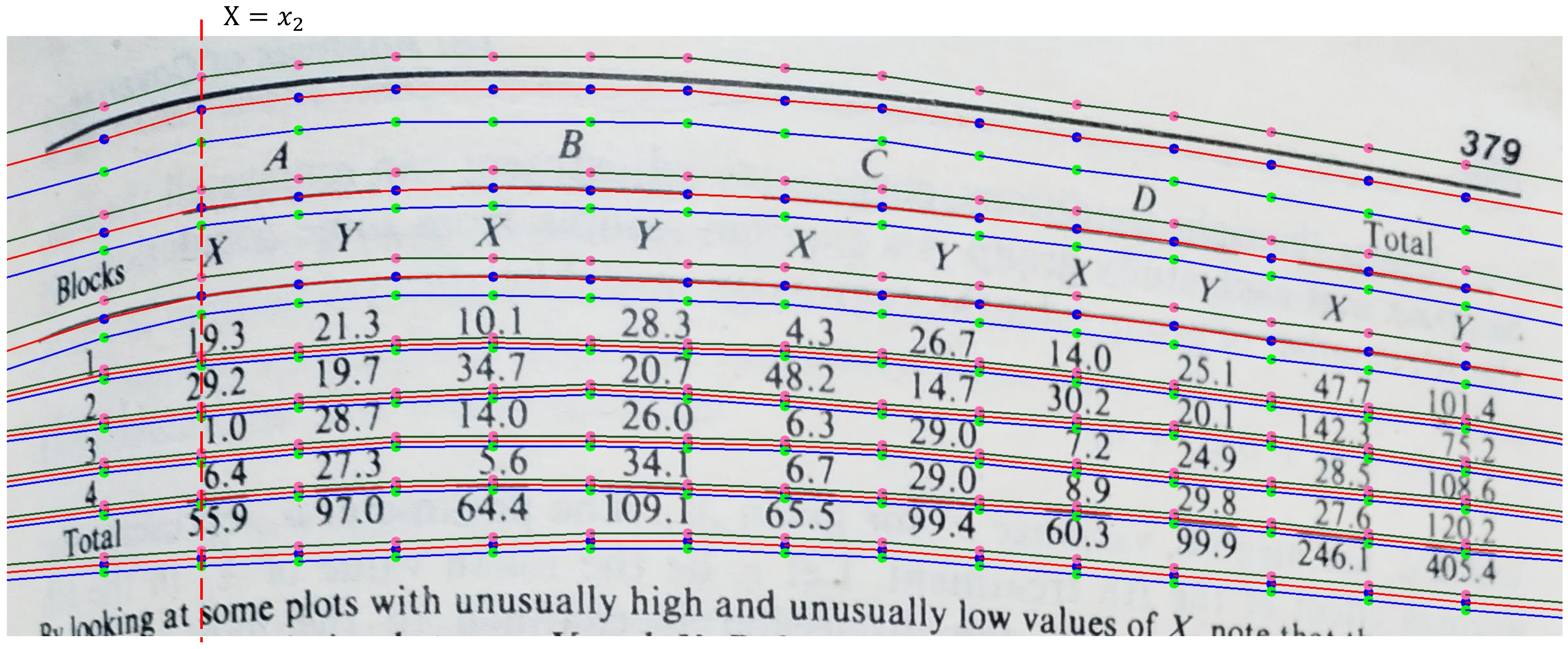}
  \caption{ An example of ground truth row separation lines.
  }
\label{fig-3}
\end{figure}

\begin{figure}[ht]
  \centering
  \includegraphics[width=1.0\linewidth]{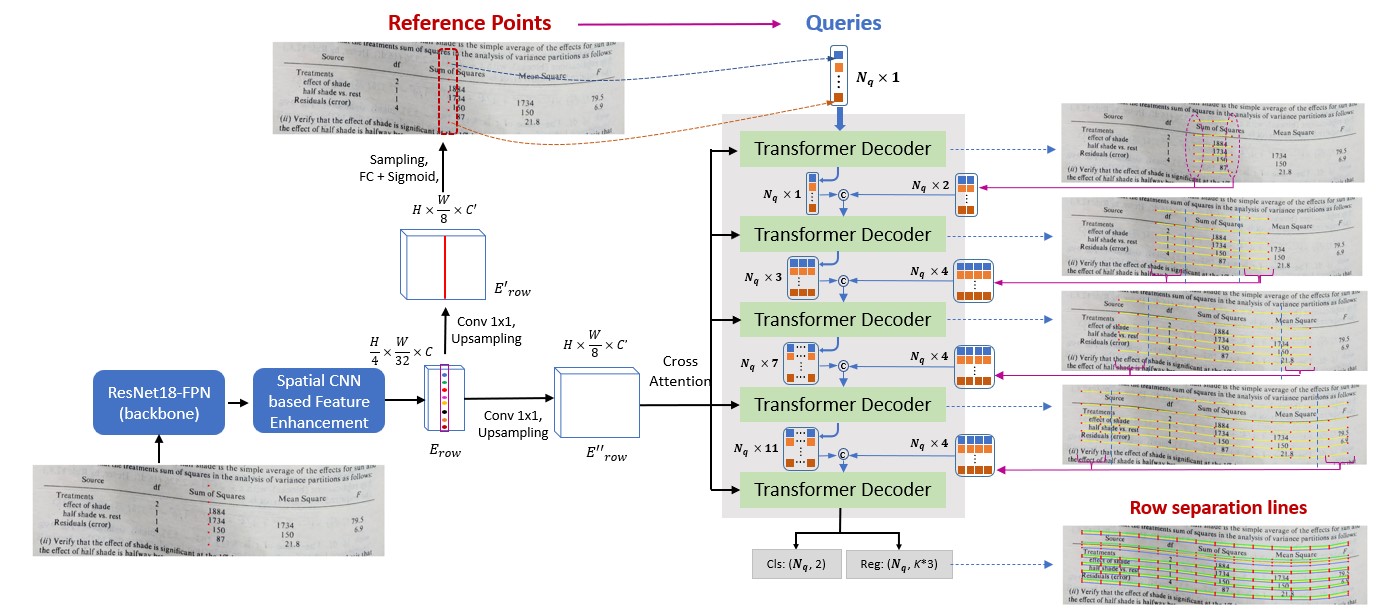}
  \caption{ The architecture of our DQ-DETR based row separation line prediction model.
  }
\label{fig-2}
\end{figure}

\subsubsection{DQ-DETR based separation line prediction}

As illustrated in Figure~\ref{fig-3}, we adopt three parallel curvilinear lines to denote the top boundary, center line, and bottom boundary of each row separator, respectively. Each curvilinear line is represented by $K=15$ points, with the x-coordinate of the $k$-th point set to $x_k=W*\frac{k}{K+1}$ ($k=1, 2, ..., K$). The DQ-DETR decoder predicts the y-coordinates of $3K$ points for each row separator. 
As depicted in Figure~\ref{fig-2}, the proposed approach begins by predicting a reference point for the center line of each row separator. These reference points serve as the initial queries of the DQ-DETR decoder, which progressively predicts the positions of other points on the center line for each row separator. Specifically, each decoder layer enhances the embeddings of the queries output by the preceding decoder layer, followed by refining the positions of their corresponding points using a regressor. Based on the refined points, one or two additional points are appended on both sides of the detected center line, serving as additional queries for the next decoder layer to detect new points of the center line for each row separator. To improve the precision of top and bottom boundary line predictions, an auxiliary task is introduced that leverages the query embeddings generated by each decoder layer to estimate the relative positions of points on these lines relative to their respective reference points on the detected center line. The resulting predicted y-coordinates of the $3K$ points for each row separator generated by the final decoder layer are deemed as the definitive outcome.
Based on the features $E_{row}$ output by the spatial CNN based feature enhancement module, the reference point detection module and the DQ-DETR decoder are attached to two different high-resolution feature maps $E_{row}'\in R^{H\times\frac{W}{8}\times C'}$ and $E_{row}''\in R^{H\times\frac{W}{8}\times C'}$ respectively, which are both generated by adding a $1\times1$ convolutional layer and an up-sampling layer sequentially to $E_{row}$.

\paragraph{\textbf{Reference point detection}}
The reference point for the center line of each row separator will be detected at a fixed position $x_\tau$ along the width direction of the raw image. Specifically, the $x_\tau^{th}$ column of $E_{row}'$ will be fed into a $1 \times 1$ convolutional layer followed by a sigmoid activation function to predict a reference point score map with the shape of $H \times 1$. Then, we apply non-maximal suppression by using a $7\times1$ max-pooling layer on the score map to suppress redundant activations for a single row separator. After that, top-100 scored row reference points are selected and further filtered by a score threshold of $0.05$. The remaining $N_q$ row reference points will be used to construct the initial queries of the following DQ-DETR decoder. Here, we set the hyper-parameter $x_\tau$ as $\lfloor \frac{W}{2} \rfloor$ for row separation line prediction and $y_\tau$ as $\lfloor \frac{H}{2} \rfloor$ for column separation line prediction in all experiments.


\paragraph{\textbf{Query initialization}} 
The first decoder layer in DQ-DETR takes all selected reference points as queries. Let $q_{j, mid}^0$ denote the initial embedding of the reference point for the $j$-th row separator. $q_{j, mid}^0$ is initialized as follows:
\begin{equation}
    \label{query_init}
    q_{j,mid}^0 = ce_{j,mid} + pe_{j,mid},
\end{equation}
where $\textbf{ce}_{j, mid}$ is a learnable content embedding and $\textbf{pe}_{j, mid}$ is a positional embedding, which is calculated by using the sinusoidal positional encoding function with the normalized coordinates of the corresponding reference point as input. In this way, we can initialize $N_q$ queries from the $N_q$ reference points. 

\begin{figure}[ht]
  \centering
  \includegraphics[height=0.6\linewidth]{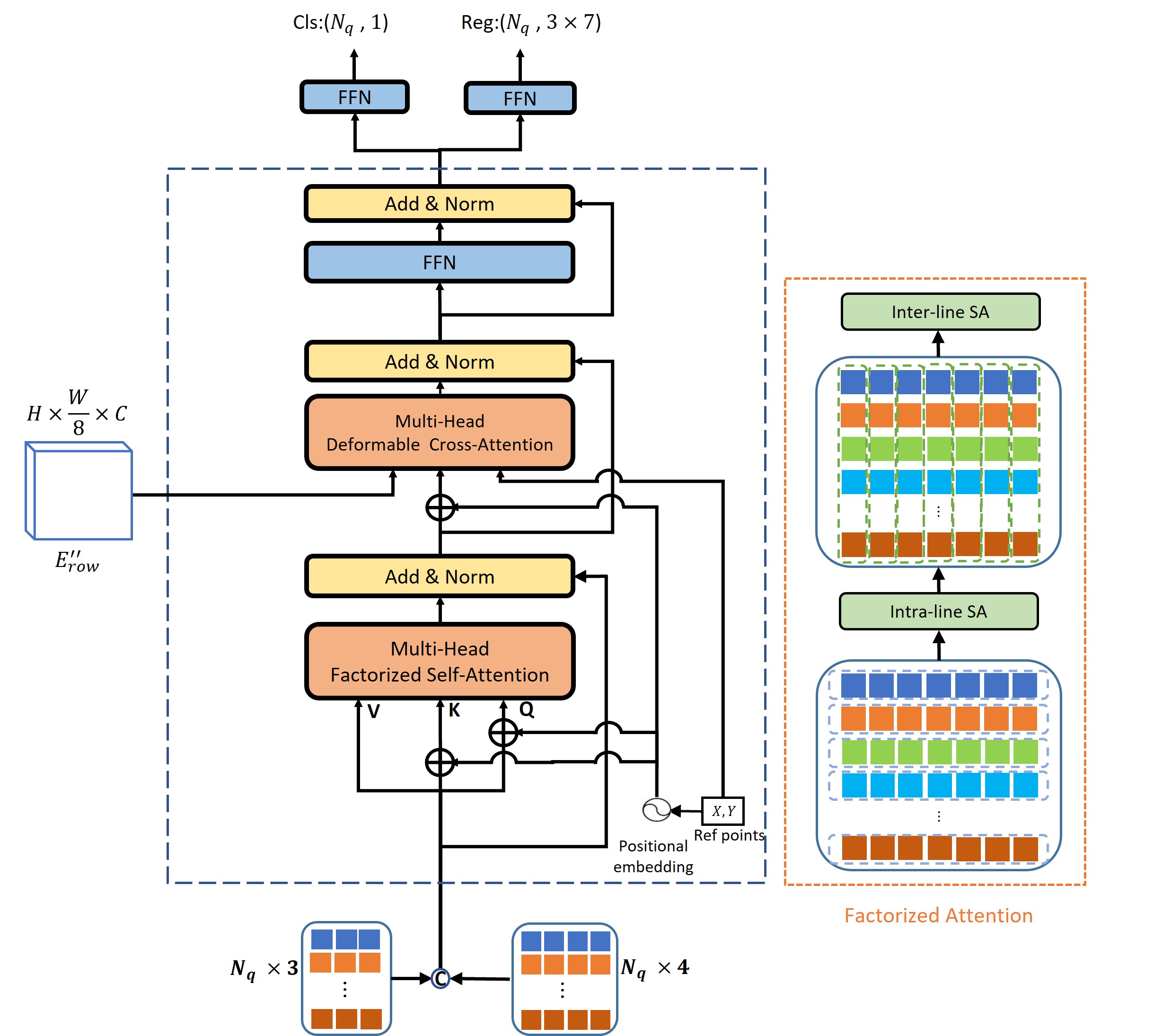}
  \caption{The architecture of DQ-DETR decoder layer for row separation line prediction.
  }
\label{fig-4}
\end{figure}

\paragraph{\textbf{Query embedding enhancement}} 
As depicted in Fig.~\ref{fig-4}, the queries input to a decoder layer form a tensor $\textbf{Q}$ with the shape of $N_q\times N_p \times D$, where $N_p$ is the number of already detected points on the center line of each separator before the current decoder layer and $D$ is the dimension of each query embedding. Here, each decoder layer is composed of a factorized self-attention (SA) module \cite{dong2021visual}, a deformable cross-attention module \cite{deformdetr2021} (CA), and an FFN. In the factorized self-attention module, instead of conducting self-attention among all queries, it first conducts intra-line self-attention and then inter-line self-attention. Specifically, all the queries belonging to the same separator will attend to each other in the intra-line self-attention and all the queries from different separators whose x-coordinates are the same will attend to each other in the inter-line self-attention.
Denote the center line of the $j$-th row separator input to the $l$-th decoder layer as an ordered point set $\{\textbf{p}_{j,k}^{l-1}|k=start,...,mid,...,end\}$, where $start=mid-\frac{N_p-1}{2}$, $mid=\frac{K+1}{2}$, $end=mid+\frac{N_p-1}{2}$, and denote the input embedding of the $k$-th point on the center line of the $j$-th row separator, i.e., $\textbf{p}_{j,k}^{l-1}$ , as $\textbf{q}_{j,k}^{l-1}$. Then, the intra-line self-attention operator $f_{intra}$ and inter-line self-attention operator $f_{inter}$ are formulated as follows: 

\noindent
\begin{align}
f_{intra}(\textbf{q}_{j,*}^{l-1}) &= \left[\bar{\textbf{q}}_{j,start}^{l-1},...,\bar{\textbf{q}}_{j,mid}^{l-1},...,\bar{\textbf{q}}_{j,end}^{l-1}\right] \notag \\
&= SA(\textbf{q}_{j,start}^{l-1},...,\textbf{q}_{j,mid}^{l-1},...,\textbf{q}_{j,end}^{l-1}), \notag \\
f_{inter}(\bar{\textbf{q}}_{*,k}^{l-1}) &= \left[\hat{\textbf{q}}_{1,k}^{l-1},\hat{\textbf{q}}_{2,k}^{l-1},...,\hat{\textbf{q}}_{N_q,k}^{l-1}\right] \notag \\
&= SA(\bar{\textbf{q}}_{1,k}^{l-1},\bar{\textbf{q}}_{2,k}^{l-1},...,\bar{\textbf{q}}_{N_q,k}^{l-1}),  \notag \\
j \in \{1,2,...,N_q&\},~~k \in \{start,...,mid,...,end\},
\end{align}
where $\hat{\textbf{q}}$ is the enhanced embedding of each query output by the factorized self-attention module. Compared with vallina self-attention, factorized self-attention can reduce the computational complexity from $O(N_q^2 N_p^2 D)$ to $O(N_q N_p^2 D + N_q^2 N_p D)$.
The updated queries are further sent into the deformable cross-attention module \cite{deformdetr2021} to aggregate high-resolution image features $E_{row}^{''}$ to leverage context information.

\begin{figure}[ht]
  \centering
  \includegraphics[width=1.0\linewidth]{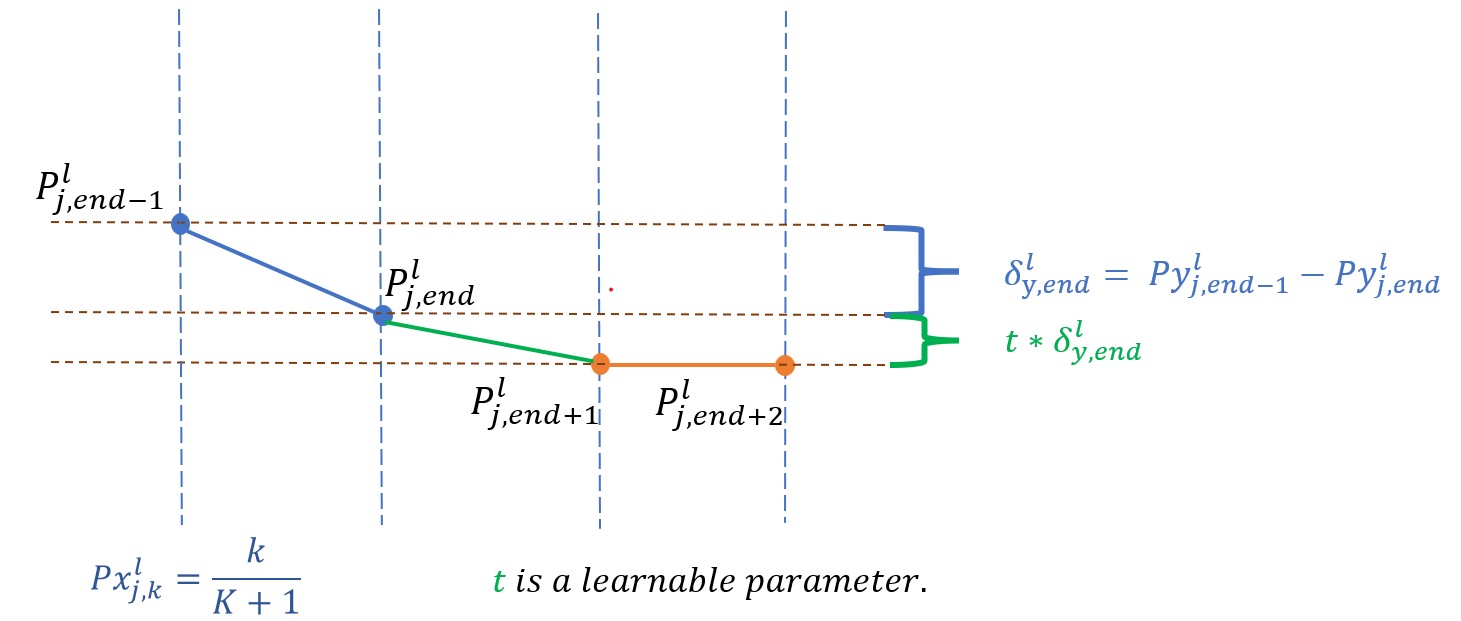}
  \caption{Illustration of the heuristic strategy utilized for generating new reference points at DQ-DETR decoder layers, with the exception of the first decoder layer.
  }
  \label{fig-extension}
\end{figure}

\paragraph{\textbf{Dynamic query generation}} 
The updated query embeddings from each decoder layer will first be fed into a regressor to refine the positions of current reference points. Given a query embedding $\textbf{q}_{j,k}^{l}$ output from the $l$-th decoder layer, we denote its current position in the input image as $\textbf{p}_{j,k}^{l-1}$ and its refined position as $\textbf{p}_{j,k}^{l}$, respectively. Since the x-coordinates of the points that need to be predicted on the center line of row separators are pre-set, only the y-coordinate $py_{j,k}^{l}$ needs to be predicted, and the x-coordinates $px_{j,k}^{l}$ can be directly obtained according to the previous definition. $\textbf{p}_{j,k}^{l}$ is calculated as follows:

\begin{equation}
    \textbf{p}_{j,k}^{l} = \left(px^l_{j,k},py^l_{j,k}\right) = \left(W*\frac{k}{K+1}, ~H*\sigma\left(\mathrm{\Delta}py_{j,k}^{l} + \sigma^{-1}(py_{j,k}^{l-1})\right)\right),
\end{equation}
where $py_{j,k}^{l-1}$ is the normalized y-coordinates of $\textbf{p}_{j,k}^{l-1}$, $\sigma(\cdot)$ is the sigmoid function and $\mathrm{\Delta}py_{j,k}^{l}$ is the predicted y-offset by the regressor.
After refining the position of each point, we add one additional point at each end of the detected center line at the first decoder layer, and two additional points at subsequent decoder layers as shown in Fig.~\ref{fig-2}. Specifically, denote a refined center-line proposal as an ordered point set $\{\textbf{p}_{j,k}^{l}|k=start,..., mid,..., end\}$, we will insert a new point $\textbf{p}_{j,start-1}^{l}$ before the first point and append a new point $\textbf{p}_{j,end+1}^{l}$ after the last point at the first layer and insert two more points $\textbf{p}_{j,start-2}^{l}$ and $\textbf{p}_{j,end+2}^{l}$ at the subsequent layers.
For the first layer, the locations of the newly added points will be simply initialized as $\textbf{p}_{j,start-1}^{l} = \textbf{p}_{j, start}^{l}$ and $\textbf{p}_{j,end+1}^{l} = \textbf{p}_{j, end}^{l}$. It is worth noting that for the first layer, $start$, $end$, and $mid$ are all equal. For the subsequent layers, the x-coordinates of the four newly added points $\textbf{p}_{j, start-1}^{l}$, $\textbf{p}_{j, start-2}^{l}$, $\textbf{p}_{j, end+1}^{l}$ and $\textbf{p}_{j, end+2}^{l}$ are pre-set and the corresponding y-coordinates will be initialized as follows:

\begin{align}
\delta_{y,start}^{l} &= py_{j, start}^{l} -py_{j, start+1}^{l}, \\
py_{j, start-2}^{l} &= py_{j, start-1}^{l} = py_{j, start}^{l} + t*\delta_{y,start}^{l}, \\
\delta_{y, end}^{l} &= py_{j, end}^{l} - py_{j, end-1}^{l}, \\
py_{j, end+2}^{l} &= py_{j, end+1}^{l} = py_{j, end}^{l} + t*\delta_{y, end}^{l},
\end{align}
where $t$ is a learnable parameter initialized as 0.5 to adjust the extension ratio. As illustrated in Fig.~\ref{fig-extension}, we take the generation of $\textbf{p}_{j, end+1}^{l}$ and $\textbf{p}_{j, end+2}^{l}$ as an example. With these extended points, we can dynamically generate $2N_q$ or $4N_q$ new queries following Eq.~\ref{query_init} and concatenate them with the existing query tensor. In this way, the center line of each row separator can be extended progressively and refined iteratively by the following decoder layers to achieve higher localization accuracy.

\paragraph{\textbf{Separation line regression}} 
The output query embeddings $\textbf{Q} \in R^{N_q \times N_p \times D}$ from each decoder layer will be fed into two feed-forward networks for classification and separation line regression, respectively. Specifically, the classifier is implemented by a fully-connected $(FC)$ layer followed by a sigmoid activation function to determine whether the corresponding reference point belongs to a separation line. If so, a regressor is used to predict the offsets of y-coordinates from the reference point of each query to the corresponding points on the center line, top boundary, and bottom boundary of the row separator, respectively. Here, the regressor is implemented by an MLP with 2 hidden layers and an output layer whose output channel dimension is 3. We consider the predicted y-coordinates of the $3K$ points for each row separator, which are produced by the last decoder layer, as the final result.

\subsubsection{Prior-enhanced bipartite matching}
Given a set of predictions and their corresponding ground-truth objects from an input image, DETR used Hungarian algorithm to assign ground-truth labels to the system predictions. However, it is found that the original bipartite matching algorithm in DETR is unstable in the training stage \cite{li2022dn}, i.e., a query could be matched with different objects in a same image in different training epochs, which slows down model convergence significantly. We find that most of the reference points detected in the first stage locate between the top and bottom boundaries of their corresponding row separators consistently in different training epochs, so we leverage this prior information to match each reference point with its closest ground-truth (GT) separator directly. In this way, the matching results will become stable during training. Specifically, we generate a cost matrix by measuring the distance between each reference point and each GT separator. If a reference point is located between the top and bottom boundaries of a GT separator, the cost is set to the distance from this reference point to the GT reference point of this separator. Otherwise, the cost is set to $INF$. Based on this cost matrix, we use the Hungarian algorithm to produce an optimal bipartite matching between reference points and ground truth separators. After getting the optimal matching result, we further remove the pair with cost $INF$ to bypass unreasonable label assignments. The experiments in Table \ref{tab:ablation3} show that the convergence of our DQ-DETR becomes much faster with our prior-enhanced bipartite matching strategy.

\begin{figure}[ht]
  \centering
  \includegraphics[width=1.0\linewidth]{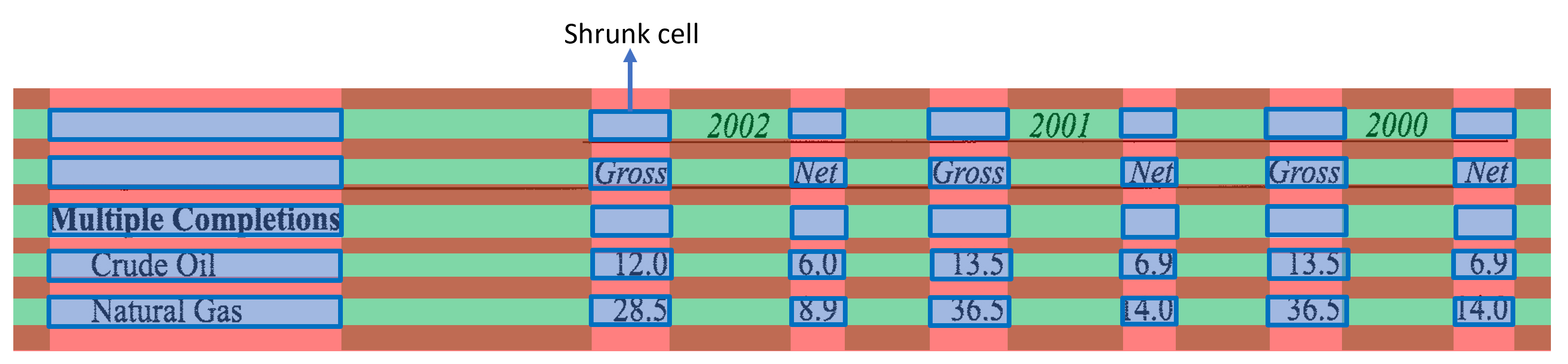}
  \caption{Examples of shrunk cells, which are the input of cell merging module.
  }
\label{fig-shrunk}
\end{figure}

\subsection{Relation network based cell merging}
\label{subsec:relation-network}
For a fair comparison with RobusTabNet \cite{ma2022robust}, we also use a lightweight relation network \cite{rn2017} to recover spanning cells. After separation line prediction, we intersect the center lines of all row and column separators to generate a grid of cells and intersect the top and bottom boundaries of all row separators with the left and right boundaries of all column separators to generate a shrunk cell box for each cell (each blue box in Fig.~\ref{fig-shrunk} represents a shrunk cell box). As shown in Fig.~\ref{fig-1}, to calculate the feature representation of each cell, we use the RoI Align algorithm \cite{maskrcnn2017} to extract a $7\times7\times C$ feature map from $P_2$ based on the bounding box of its shrunk cell box and feed this feature map into a two-layer MLP with 512 nodes at each layer to generate a 512-d feature vector first. These feature vectors can be arranged in a grid with $N$ rows and $M$ columns to form a feature map $F_{cell}\in R^{N\times M\times512}$, which is then enhanced by three repeated feature enhancement blocks to generate an enhanced feature map. Each feature enhancement block contains three parallel branches with a row-wise max-pooling layer, a column-wise max-pooling layer, and a $3\times3$ convolutional layer, respectively. The output feature maps of these three branches are concatenated together and convoluted by a $1\times1$ convolutional layer for dimension reduction. Finally, for each pair of adjacent cells, we concatenate their feature vectors extracted from the enhanced feature map and an 18-d spatial compatibility feature vector introduced in \cite{rn2017} to generate a new feature vector, which is fed into a binary classifier to predict whether these two cells should be merged or not. The binary classifier is implemented with a 2-hidden-layer MLP with 512 nodes at each hidden layer and a sigmoid activation function.

\section{Loss function}

The loss functions for training the split module and the cell merging module in TSRFormer are defined in this section. For the split module, we take row separator prediction as an example and denote the corresponding loss items as $L_{*}^{row}$. Likewise, we can also calculate the losses for column separator prediction, denoted as $L_{*}^{col}$.

\textbf{Reference point detection.} We adopt a variant of focal loss \cite{focal2017} to train the row reference point detection module:
\begin{equation}
L_{ref}^{row}=-\frac{1}{N_r}\sum_{i=1}^H \left\{
    \begin{array}{ll}
        (1-p_{i})^\alpha log(p_{i}),  &   p^*_{i}=1 \\
        (1-p^*_{i})^\beta p_{i}^\alpha log(1-p_{i}),  &   otherwise
    \end{array}
\right.
\end{equation}

\noindent where $N_r$ is the number of row separation lines, $\alpha$ and $\beta$ are two hyper-parameters set to 2 and 4 respectively as in \cite{cornernet2018}, $p_i$ and $p_i^*$ are the predicted and ground-truth labels for the $i^{th}$ pixel in the $x_\tau^{th}$ column of $E_{row}'$. Here, $p_i^*$ has been augmented with unnormalized Gaussians, which are truncated at the boundary of separators, to reduce the penalty around the ground-truth reference point locations. Specifically, let $(y_k,x_\tau)$ denote the ground-truth reference point for the $k^{th}$ row separator, which is the intersection point of the center line of this row separator and the vertical line $x = x_\tau$. The vertical distance between the top and bottom boundaries of the $k^{th}$ row separator is taken as its thickness, denoted as $w_k$. Then, $p_i^{*}$ can be defined as follows:
\begin{equation}
p_i^{*}=\left\{
    \begin{array}{ll}
        exp(-\frac{(i-y_k)^2}{2\sigma_k^2}),  &   if~i \in (y_k - \frac{w_k}{2}, y_k + \frac{w_k}{2}) \\
        0,  &   otherwise
    \end{array}
\right.
\end{equation}

\noindent where $\sigma_k=\sqrt{\frac{w_k^2}{2ln(10)}}$ is adaptive to the thickness of the separator to make sure that $p_i^{*}$ within this row separator is no less than 0.1. 

\textbf{Separation line regression.} Let $y=\{(c_i,l_i)|i=1,...,M\}$ denote the set of ground-truth row separators, where $c_i$ and $l_i$ indicate the target class and row separator position respectively, $y^*=\{(c_q^*,l_q^*)|q=1,...,Q\}$ denote the set of predictions. After getting the optimal bipartite matching result $\hat{\sigma}$, the loss of row separation line regression can be calculated as:
\begin{equation}
    L_{line}^{row}=\sum_{i=1}^Q[L_{cls}(c_i, c_{\hat{\sigma}(i)}^*)+\pmb{1}_{\{c_i\neq \varnothing\}}L_{reg}(l_i,l_{\hat{\sigma}(i)}^*)]
\end{equation}

\noindent where $L_{cls}$ is focal loss and $L_{reg}$ is L1 loss. For each decoder layer, the loss of regression $L_{line}^{row}$ is added to assist training. Due to the dynamic change in the number of predicted points at different decoder layers, the set of ground-truth row separators $y$ needs to extract the corresponding number of points for each separator, and the same loss function is performed. The overall loss of row separation line regression is as follows:
\begin{equation}
    L_{line}^{row,all}=\sum_{l=1}^{L} L_{line}^{row,l} = \sum_{l=1}^{L}\sum_{i=1}^Q[L_{cls}(c_i^l, c_{\hat{\sigma}(i)}^{*,l})+\pmb{1}_{\{c_i^l\neq \varnothing\}}L_{reg}(l_i^l,l_{\hat{\sigma}(i)}^{*,l})]
\end{equation}
where $L$ means the number of the decoder layers, $(c^l, l^l)$ and $(c^{*,l}, l^{*,l})$ represent the corresponding part of ground-truth row separators and prediction at the $l$-th layer, respectively.

\textbf{Cell merging.} The loss $L_{merge}$ of the cell merging module is a binary cross-entropy loss:
\begin{equation}
    L_{merge}=\frac{1}{|S_{rel}|}\sum_{i\in S_{rel}} BCE(P_i, P_i^*)
\end{equation}

\noindent where $S_{rel}$ denotes the set of sampled cell pairs,
$P_i$ and $P_i^*$ denote the predicted and ground-truth labels for the $i^{th}$ cell pair, respectively.

\textbf{Overall loss.} All the modules in TSRFormer can be trained jointly. The overall loss function is as follows:

\begin{align}
    L=\lambda (L_{ref}^{row}+L_{ref}^{col})+L_{line}^{row, all}+L_{line}^{col, all}+L_{merge}
\end{align}

\noindent where $\lambda$ is a control parameter set to 0.2 in our experiments.

\section{Experiments}
\subsection{Datasets and evaluation protocols}
 We conduct experiments on four popular public benchmarks, including SciTSR \cite{chi2019complicated}, PubTabNet \cite{zhong2020image}, FinTabNet \cite{zheng2020global} and WTW \cite{long2021parsing}, to verify the effectiveness of the proposed method. Moreover, we also collected a more challenging in-house dataset, which includes many challenging tables with complex structures, borderless cells, large blank spaces, empty or spanning cells as well as distorted or even curved shapes, to demonstrate the superiority of our TSRFormer.

\textbf{SciTSR} \cite{chi2019complicated} contains 12,000 training samples and 3,000 testing samples of axis-aligned tables cropped from scientific literatures. There are also 716 complicated tables selected by authors from the testing set to create a more challenging test subset, called SciTSR-COMP. 
In this dataset, the cell adjacency relationship metric \cite{gobel2013icdar} is used as the evaluation metric.
Instead of comparing the IoU of bounding boxes of detected cells (or contents) and the ground-truth cells (or contents), whether the text contents in the detected cell and the ground-truth cell match exactly is used in the evaluation tool. We assign each candidate text content from the PDF files of SciTSR, whose bounding boxes is denoted as $b_{text}$, into a detected cell box $b_{det}$ if the following condition is satisfied, i.e.,
\begin{equation}
    \frac{Area(b_{text} \cap b_{det})}{Area(b_{text})} > 0.5
\end{equation}
where $Area(b_{text})$ and $Area(b_{text} \cap b_{det})$ denote the area of $b_{text}$ and the area of the overlap between $b_{text}$ and $b_{det}$, respectively. Additionally, we take empty cells into account when evaluating.

\textbf{PubTabNet} \cite{zhong2020image} contains 500,777 training, 9,115 validation, and 9,138 testing images generated by matching the XML and PDF representations of scientific articles. All the tables are axis-aligned. Since the annotations of the testing set are not released, we only report results on the validation set. This work proposed a new Tree-Edit-Distance-based Similarity (TEDS) metric for table recognition task, which can identify both table structure recognition and OCR errors. However, taking OCR errors into account may cause an unfair comparison because of the different OCR models used by different TSR methods. Some recent works \cite{GTE2021,TabStruct2020,Qiao2021LGPMACT} have proposed a modified TEDS metric named TEDS-Struct to evaluate table structure recognition accuracy only by ignoring OCR errors. We also use this modified metric to evaluate our approach on this dataset.

\textbf{FinTabNet} \cite{zheng2020global} is a large dataset containing more than 70K pages with full table bounding boxes and structure annotations (train/val/test=
61801/7191/7085) and more than 110k axis-aligned tables with cell bounding boxes (train/val/test=91596/10635/10656) from the annual reports of the S\&P 500 companies. We use the 110k cropped table images to evaluate our approach. Following the original FinTabNet paper \cite{zheng2020global}, the TEDS-Struct metric is used as the evaluation metric.

\textbf{WTW} \cite{long2021parsing} contains 10,970 training images and 3,611 testing images collected from wild complex scenes. This dataset focuses on bordered tabular objects only and contains the annotated information of table id, tabular cell coordinates, and row/column information. We crop table regions from original images for both training and testing and follow \cite{long2021parsing} to use the cell adjacency relationship (IoU=0.6) \cite{gobel2012methodology} as the evaluation metric of this dataset. 

\textbf{cTDaR TrackB2-Modern} \cite{gao2019icdar} contains no images for training, but 100 images with annotations are provided as testing data. RobusTabNet \cite{ma2022robust} manually labelled the structures of tables in the cTDaR TrackA modern subset, which contains 600 training images. It has been checked that there is no overlap between the 600 training images and the 100 testing images. To evaluate our approach on this dataset, we follow RobusTabNet \cite{ma2022robust} to train our model on that dataset. The cTDaR TrackB metric \cite{gao2019icdar} is used as the evaluation metric of this dataset. During the evaluation, the convex hull of the content is used to represent a cell. Note that both table region detection and table structure recognition have to be done on this dataset. To validate the effectiveness of our table structure recognition module, we use the same table detector as RobusTabNet \cite{ma2022robust}. Following previous works \cite{zou2020deep, prasad2020cascadetabnet}, the experimental results reported by us are based on the average of IoU=0.6, 0.7, 0.8, 0.9.

\textbf{In-House dataset} contains 40,590 training images and 1,053 testing images, cropped from heterogeneous document images including scientific publications, financial statements, invoices, etc. Most images in this dataset are captured by cameras so tables in these images may be skewed or even curved. Some examples can be found in Fig. \ref{fig-comp}, Fig. \ref{fig-8}, Fig. \ref{fig-5} and Fig. \ref{fig-7}. The same adjacency relation-based metric as cTDaR TrackB is used for evaluation. We use ground-truth text boxes as table contents and report results based on IoU=0.9.

It is worth noting that the definitions of Cell Adjacency Relationship used in the respective evaluation metrics of SciTSR, WTW, cTDaR TrackB2-Modern, and in-house dataset are the same and all based on \cite{gobel2012methodology}.


\subsection{Ground-truth generation}
We take the row separation line prediction branch as an example to introduce how to generate the ground-truth top boundary, center line, and bottom boundary of each row separator. In SciTSR, PubTabNet, and FinTabNet datasets, the row and column spanning information of each cell in each table are annotated to represent the table structure, and the bounding boxes of text-lines in each cell are annotated to represent the content of each cell. As tables in these datasets are axis-aligned, we calculate a minimum bounding box for each table row by using an axis-aligned rectangle to enclose the text-line boxes in all non-spanning cells in this row with the minimum area. The top and bottom boundaries of the minimum bounding box of each table row are taken as the ground-truth bottom and top boundaries of the row separators above and below this table row, respectively. In the WTW dataset, since only the borders of bordered cells are labeled, we generate the ground-truth center lines of row separators above and below a table row by connecting and extending the top and bottom borders of annotated cell boxes in this row, respectively, and set the width of all the separator masks to 8 pixels to obtain the top and bottom boundaries of each row separator. As tables in the in-house dataset could be distorted, in addition to the row and column spanning information of each cell, the borders of both table cells and text-lines are also annotated. We calculate the labeled row separation lines by connecting the top and bottom borders of cell boxes in each table row first. Given these labeled row separation lines as well as the bounding boxes of text-lines in each cell, we follow \cite{ma2022robust} to generate a segmentation mask for each row separator by moving the corresponding separation line upwards and downwards respectively until it touches a text box that belongs to a non-spanning cell. Then, the top boundary, center line, and bottom boundary of each row separator mask are considered as the ground-truth lines of each row separator. 

We use cells detected by the split model to generate positive and negative relational pairs for training the cell merging module. Given detected/ground-truth row and column separators from a table image, we intersect their center lines and boundaries to generate cell boxes and shrunk cell boxes for detected/ground-truth cells, respectively. Then, we assign each detected cell $b_{det}$, whose shrunk cell box is denoted as $b_{det\_shrunk}$, to a ground-truth cell box $b_{gt}$ if the following condition is satisfied, i.e.,
\begin{equation}
    \frac{Area(b_{det\_shrunk} \cap b_{gt})}{Area(b_{det\_shrunk})} > 0.5
\end{equation}
where $Area(b_{det\_shrunk})$ and $Area(b_{det\_shrunk} \cap b_{gt})$ denote the area of $b_{det\_shrunk}$ and the area of the overlap between $b_{det\_shrunk}$ and $b_{gt}$, respectively. After that, each detected cell is paired with each of its 4-connected cells to construct candidate relational pairs. If two cells in a relational pair are assigned to the same ground-truth cell box, we give this relational pair a positive label, otherwise a negative label. During training, we ignore all the negative relational pairs that contain cells not assigned to any ground-truth cell. 

\subsection{Implementation details}
All experiments are implemented in Pytorch v1.8.1 and conducted on a workstation with 8 Nvidia Tesla V100 GPUs. We use ResNet18-FPN as the backbone and set the channel number of $P_2$ to 64 in all experiments. The weights of RestNet-18 are initialized with a pre-trained model for the ImageNet classification task. The models are optimized by AdamW \cite{loshchilov2017decoupled} algorithm with batch size 16. We use a polynomial decay schedule with the power of 0.9 to decay the learning rate, and the initial learning rate, betas, epsilon, and weight decay are set as
1e-4, (0.9, 0.999), 1e-8 and 5e-4, respectively. Synchronized BatchNorm is applied during training. In DQ-DETR based split modules, we set the channel number of $E_{row}'$/$E_{col}'$ to 256, and the query dimension, head number, and dimension of feedforward networks in transformer decoder layers to 256, 16 and 1024, respectively.

In the training phase, we randomly rescale the shorter side of table images to a number in \{416, 512, 608, 704, 800\} while keeping the aspect ratio for all datasets except WTW. For WTW, we follow \cite{long2021parsing} to resize both sides of each training image to 1024 pixels. In each image, we sample a mini-batch of 64 hard positive and 64 hard negative cell pairs for the cell merging module. The hard samples are selected with the OHEM \cite{shrivastava2016ohem} algorithm. During training, we first train the reference point detection module for $N$ epochs and then jointly train this module and the separation line regression module for $N$ epochs. Finally, the cell merging module is further added and jointly trained for another $N$ epochs. Here, $N$ is set as 12 for PubTabNet and 20 for the other datasets. 

In the testing phase, we rescale the longer side of each image to 1024 while keeping the aspect ratio for SciTSR, PubTabNet, FinTabNet, and in-house dataset. For WTW, the strategy is the same as in training. 

\setlength{\tabcolsep}{4pt}
\begin{table}[ht]
\setlength{\belowcaptionskip}{0.2cm}
\small
\begin{center}
\caption{Results on SciTSR dataset.}
\label{table:SciTSR}
\begin{tabular}{ccccccc}
\hline
\multirow{2}{*}{Methods} & \multicolumn{3}{c}{SciTSR} & \multicolumn{3}{c}{SciTSR-COMP}\\
\cmidrule(lr){2-4} \cmidrule(lr){5-7} & Prec. (\%) & Rec. (\%) & F1. (\%) & Prec. (\%) & Rec. (\%) & F1. (\%)\\
\hline
TabStruct-Net \cite{TabStruct2020}  & 92.7 & 91.3 & 92.0 & 90.9 & 88.2 & 89.5\\
GraphTSR \cite{chi2019complicated}  & 95.9 & 94.8 & 95.3 & 96.4 & 94.5 & 95.5\\
LGPMA \cite{Qiao2021LGPMACT}  & 98.2 & 99.3 & 98.8 & 97.3 & 98.7 & 98.0\\
FLAG-Net \cite{FLAG2021}  & 99.7 & 99.3 & \textbf{99.5} & 98.4 & 98.6 & 98.5\\
RobusTabNet\cite{ma2022robust} & 99.4 & 99.1 & 99.3 & 99.0 & 98.4 & 98.7 \\
\hline
TSRFormer w/ DQ-DETR  & 99.5 & 99.3 & 99.4 & 99.1 & 98.6 & \textbf{98.9}\\
\hline
\end{tabular}
\end{center}
\end{table}
\setlength{\tabcolsep}{1.4pt}

\setlength{\tabcolsep}{4pt}
\setlength{\belowcaptionskip}{0.2cm}
\begin{table}[ht]
\small
\begin{center}
\caption{Results on PubTabNet dataset.}
\label{table:PubTabNet}
\begin{tabular}{cccc}
\hline\noalign{\smallskip}
Methods & Training Dataset & TEDS (\%) & TEDS-Struct (\%)\\
\noalign{\smallskip}
\hline
\noalign{\smallskip}
EDD \cite{zhong2020image}  & PubTabNet & 88.3 & -\\
TableStruct-Net \cite{TabStruct2020}  & SciTSR & - & 90.1\\
GTE \cite{GTE2021}  & PubTabNet & - & 93.0\\
LGPMA \cite{Qiao2021LGPMACT}  & PubTabNet & 94.6 & 96.7\\
FLAG-Net \cite{FLAG2021}  & SciTSR & 95.1 & -\\
RobusTabNet \cite{ma2022robust} & PubTabNet & - & 97.0 \\
\hline
TSRFormer w/ DQ-DETR  & PubTabNet & - & \textbf{97.5}\\
\hline
\end{tabular}
\end{center}
\end{table}
\setlength{\tabcolsep}{1.4pt}

\subsection{Comparisons with prior arts}
\begin{figure*}[ht]
    \centering
    \includegraphics[width=1.0\linewidth]{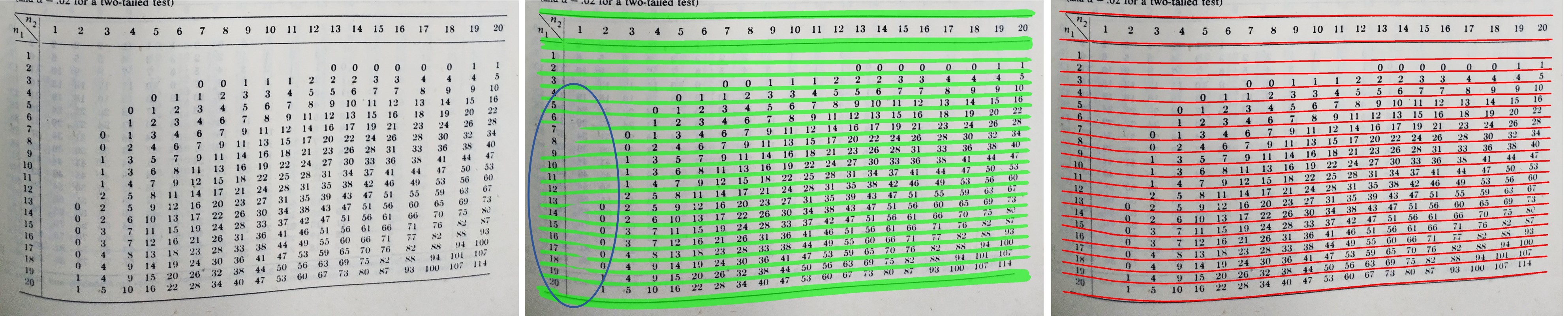}
    \caption{Qualitative results of RobusTabNet (middle) and our proposed DQ-DETR (right) for row separation line prediction on a challenging curved table with borderless cells and large blank spaces.}
    \label{fig-8}
\end{figure*}

We compare our DQ-DETR based TSRFormer with previous state-of-the-art TSR methods on five public datasets first, including SciTSR, PubTabNet, FinTabNet, WTW, and cTDaR TrackB2-Modern. 
As reported in Table \ref{table:SciTSR}, our approach achieves comparable performance on both the full testing set and the SciTSR-COMP subset (containing complicated tables only) against previous best methods. Moreover, the smaller accuracy degradation of our approach on the SciTSR-COMP subset demonstrates that our approach is more robust to tables with complex structures than other TSR methods. On the competitive PubTabNet dataset (see Table \ref{table:PubTabNet}), our approach achieves the highest TEDS-Struct score of 97.5\%. On FinTabNet (see Table \ref{table:FinTabNet}), our approach outperforms TableFormer \cite{nassar2022tableformer} by 1.6\% absolutely in terms of TEDS-Struct score. On the more challenging WTW dataset (see Table \ref{tab:WTW}), our DQ-DETR based TSRFormer achieves 1.9\% higher F1-score than Cycle-CenterNet \cite{long2021parsing}, which is specially designed for recognizing distorted bordered tables. 
On cTDaR TrackB2-Modern, the table detector in RobusTabNet\cite{ma2022robust} and our table structure recognizer are combined together to conduct end-to-end evaluation. Since the outputs of our approach are cell boxes rather than convex hulls of cell contents, for the sake of fair comparison, we use the same text detection algorithm as CascadeTabNet \cite{prasad2020cascadetabnet} to detect texts in each image and then assign them to table cells if 80\% of a text box is located in a cell box. 
As shown in Table~\ref{tab:cTDaR_TrackB2_Modern}, our DQ-DETR based TSRFormer achieves 0.6\% higher F1-score than RobusTabNet \cite{ma2022robust}. It is noted that the score on this dataset is relatively low mainly because the localization accuracy of text boxes generated by OCR is not good enough.
To further verify the robustness of our DQ-DETR based TSRFormer to different types of distorted tables, we compare it to RobusTabNet\cite{ma2022robust}, which is one of the previous best performing TSR methods, on our challenging in-house dataset. As shown in Table \ref{tab:InHouse-05}, DQ-DETR based TSRFormer outperforms RobusTabNet significantly by improving the F1-score from 92.3\% to 95.7\% under the default evaluation setting. The qualitative comparison examples in Fig.~\ref{fig-8} show that our approach is more robust to some challenging cases, like distorted tables with many empty cells, than RobusTabNet. 

\begin{figure*}[t]
    \centering
    \includegraphics[width=1.0\linewidth]{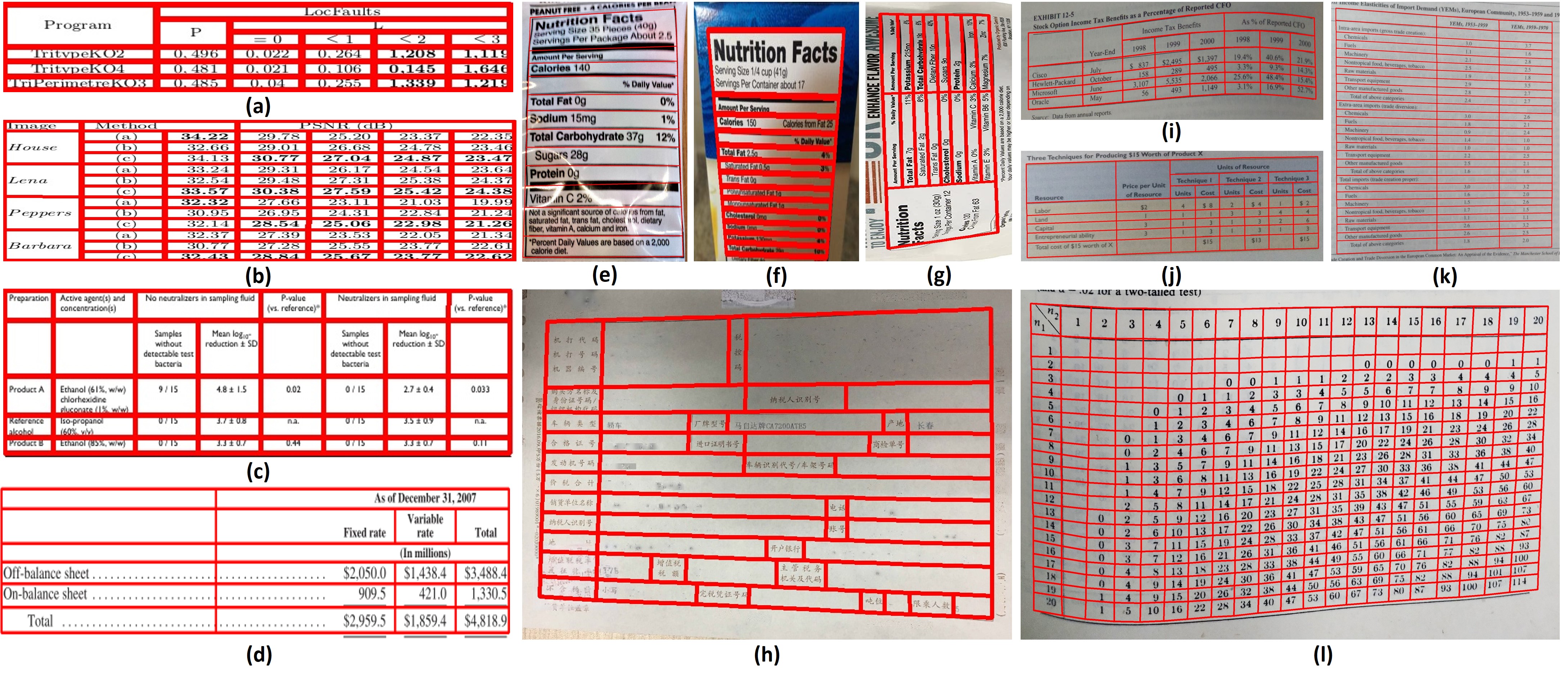}
    \caption{Qualitative results of our approach. (a-b) are from SciTSR, (c) is from PubTabNet, (d) is from FinTabNet, (e-h) are from WTW, (i-l) are from the in-house dataset.}
    \label{fig-5}
\end{figure*}

More qualitative results of our approach on different datasets are shown in Fig.~\ref{fig-5}, from which we can observe that our approach is robust to tables with complex structures, borderless cells, large blank spaces, empty or spanning cells as well as distorted or even curved shapes.

\setlength{\tabcolsep}{4pt}
\begin{table}[!ht]
\setlength{\belowcaptionskip}{0.2cm}
\small
\begin{center}
\caption{Results on FinTabNet dataset.}
\label{table:FinTabNet}
\begin{tabular}{cccc}
\hline\noalign{\smallskip}
Methods & Training Dataset & TEDS-Struct (\%)\\
\noalign{\smallskip}
\hline
\noalign{\smallskip}
Det-Base \cite{GTE2021} & PubTabNet & 41.6\\
EDD \cite{zhong2020image} & PubTabNet & 90.6\\
GTE \cite{GTE2021}  & PubTabNet \& FinTabNet & 91.0\\
TableFormer \cite{nassar2022tableformer} & FinTabNet & 96.8\\
\hline
TSRFormer w/ DQ-DETR  & FinTabNet & \textbf{98.4}\\
\hline
\end{tabular}
\end{center}
\end{table}
\setlength{\tabcolsep}{1.4pt}

\begin{table}[ht]
    \setlength{\belowcaptionskip}{0.2cm}
    \setlength{\tabcolsep}{1.5mm}
    \footnotesize
    \centering
    \caption{Results on WTW dataset.}
    \label{tab:WTW}
    \begin{tabular}{cccc}
        \hline\noalign{\smallskip}
        Methods & Prec. (\%) & Rec. (\%) & F1. (\%)\\
        \noalign{\smallskip}
        \hline
        \noalign{\smallskip}
        Cycle-CenterNet \cite{long2021parsing}  & 93.3 & 91.5 & 92.4\\
        \hline
        TSRFormer w/ DQ-DETR  & \textbf{94.5} & \textbf{94.0} & \textbf{94.3}\\
        \hline
    \end{tabular}
\end{table}

\begin{table}[h!]
    \setlength{\tabcolsep}{2.9pt}
    \footnotesize
    \centering
    \caption{TSR Performance comparison on ICDAR2019 cTDaR TrackB2-Modern. * indicates that the results are from \cite{gao2019icdar}.}
    \label{tab:cTDaR_TrackB2_Modern}
    \begin{tabular}{ c  c  c  c  c  c  c  c  c  c  c  c  c  c  c  c  c}
        \toprule
        \multirow{2}{*}{Methods} & \multicolumn{3}{c}{IoU@0.6(\%)} && \multicolumn{3}{c}{IoU@0.7(\%)} &&
        \multicolumn{3}{c}{IoU@0.8(\%)} &&
        \multicolumn{3}{c}{IoU@0.9(\%)} & WAvg. \\\cline{2-4}\cline{6-8}\cline{10-12}\cline{14-16}
         & P & R & F1 && P & R & F1 && P & R & F1 && P & R & F1 & F1(\%)\\
        \midrule
        Zou et al.\cite{zou2020deep} & 18.8 & 10.1 & 13.1 && - & - & - && 1.7 & 0.9 & 1.2 && - & - & - & - \\
        NLPR-PAL* & 32.2 & 42.1 & 36.5 && 26.9 & 35.1 & 30.5 && 17.2 & 22.5 & 19.5 && 3.1 & 4.0 & 3.5 & 20.6 \\
        CascadeTabNet\cite{prasad2020cascadetabnet} & 49.9 & 39.0 & 43.8 && 40.3 & 31.5 & 35.4 && 21.6 & 16.9 & 19.0 && 4.1 & 3.2 & 3.6 & 23.2 \\
        GTE\cite{zheng2020global} & - & - & 38.5 && - & - & - && - & - & - && - & - & - & 24.8 \\
        RobusTabNet\cite{zheng2020global} & 76.4 & 76.8 & 76.6 && 71.3 & 71.6 & 71.4 && 58.1 & 58.4 & 58.3 && 25.7 & 25.8 & 25.8 & 55.3 \\
        \midrule
        TSRFormer w/ DQ-DETR & 74.2 & 72.1 & 73.2 && 71.4 & 69.4 & 70.4 && 62.0 & 60.2 & 61.1 && 29.1 & 28.2 & 28.6 & \textbf{55.9} \\
        \bottomrule
    \end{tabular}
\end{table}

\begin{table}[!h]
\small
\setlength{\belowcaptionskip}{0.2cm}
\setlength{\tabcolsep}{1mm} 
\footnotesize
\centering
\caption{Results on In-house dataset.}
\label{tab:InHouse-05}
\begin{tabular}{cccc}
    \hline
   Methods & Prec. (\%) & Rec. (\%) & F1. (\%) \\
    \noalign{\smallskip}
    \hline
    \noalign{\smallskip}
    SPLERGE \cite{SPLERGE}  & 85.4 & 82.3 & 83.8 \\
    RobusTabNet \cite{ma2022robust} & 93.0 &91.6 &92.3\\
    TSRFormer w/ SepRETR & \textbf{95.1} & \textbf{95.3} & \textbf{95.2} \\
    TSRFormer w/ DQ-DETR & \textbf{95.7} & \textbf{95.7} & \textbf{95.7} \\
    \hline
\end{tabular}

\end{table}

\begin{table}[!h]
    \setlength{\belowcaptionskip}{0.2cm}
    \setlength{\tabcolsep}{1.5mm}
    \footnotesize
    \centering
    \caption{Results on In-house dataset. Limitation threshold denotes the limitation while assigning text-lines to predicted cells. The default limitation threshold is 50\%. The limitation of 80\% is more strict.}
    \label{tab:Inhouse}
    \begin{tabular}{c|ccc|ccc}
        \hline
        \multirow{2}{*}{Methods} & \multicolumn{3}{c|}{Limitation threshold = 50\%}                                                      & \multicolumn{3}{c}{Limitation threshold = 80\%}                                                     \\ \cline{2-7} 
                                 & \multicolumn{1}{l}{Prec. (\%)} & \multicolumn{1}{l}{Rec. (\%)} & \multicolumn{1}{l|}{F1. (\%)} & \multicolumn{1}{l}{Prec. (\%)} & \multicolumn{1}{l}{Rec. (\%)} & \multicolumn{1}{l}{F1. (\%)} \\ \hline
        SPLERGE                  & 85.4                          & 82.3                         & 83.8                         & 80.6                          & 67.6                         & 73.6 (-10.2)                       \\ \hline
        RobusTabNet              & 93.0                          & 91.6                         & 92.3                         & 92.2                          & 89.6                         & 90.9 (-1.4)                       \\ \hline
        TSRFormer w/ SepRETR     & 95.1                          & 95.3                         & 95.2                         & 94.4                          & 93.8                         & 94.1 (-1.1)                       \\ \hline
        TSRFormer w/ DQ-DETR     & 95.7                          & 95.7                         & \textbf{95.7 }                        & 95.3                          & 94.8                         & \textbf{95.1 (-0.6)}                       \\ \hline
    \end{tabular}
\end{table}

\begin{table}[ht]
\small
\setlength{\belowcaptionskip}{0.2cm}
\setlength{\tabcolsep}{1mm} 
\footnotesize
\centering
\caption{Comparisons of TSRFormer w/ SepRETR and TSRFormer w/ DQ-DETR on different datasets.}
\label{tab:sep-dq}
\begin{tabular}{cccccc}
    \hline
   Methods & Dataset & Prec. (\%) & Rec. (\%) & F1. (\%) & TEDS-Struct (\%)\\
    \noalign{\smallskip}
    \hline
    \noalign{\smallskip}
    TSRFormer w/ SepRETR & SciTSR & \textbf{99.5} & \textbf{99.4} & \textbf{99.4} & -\\
    TSRFormer w/ DQ-DETR & SciTSR & \textbf{99.5} & 99.3 & \textbf{99.4} & -\\
    \hline
    TSRFormer w/ SepRETR & SciTSR-COMP & \textbf{99.1} & \textbf{98.7} & \textbf{98.9} & -\\
    TSRFormer w/ DQ-DETR & SciTSR-COMP & \textbf{99.1} & 98.6 & \textbf{98.9} & -\\
    \hline
    TSRFormer w/ SepRETR & PubTabNet & - & - & - & \textbf{97.5}\\
    TSRFormer w/ DQ-DETR & PubTabNet & - & - & - & \textbf{97.5}\\
    \hline
    TSRFormer w/ SepRETR & FinTabNet & - & - & - & 98.3 \\
    TSRFormer w/ DQ-DETR & FinTabNet & - & - & - & \textbf{98.4} \\
    \hline
    TSRFormer w/ SepRETR  & WTW & 93.7 & 93.2 & 93.4\\
    TSRFormer w/ DQ-DETR  & WTW & \textbf{94.5} & \textbf{94.0} & \textbf{94.3}\\
    \hline
    TSRFormer w/ SepRETR  & cTDaR TrackB2-Modern & - & - & 53.6 (WAvg.)\\
    TSRFormer w/ DQ-DETR  & cTDaR TrackB2-Modern & - & - &\textbf{ 55.9} (WAvg.)\\    
    \hline
    TSRFormer w/ SepRETR & In-house & 95.1 & 95.3 & 95.2 & -\\
    TSRFormer w/ DQ-DETR & In-house & \textbf{95.7} & \textbf{95.7} & \textbf{95.7} & -\\
    \hline
\end{tabular}

\end{table}

\subsection{Ablation studies}
We conduct a series of experiments to evaluate the effectiveness of different modules in our approach on our in-house dataset.

\begin{figure}
    \centering
    \includegraphics[width=1.0\linewidth]{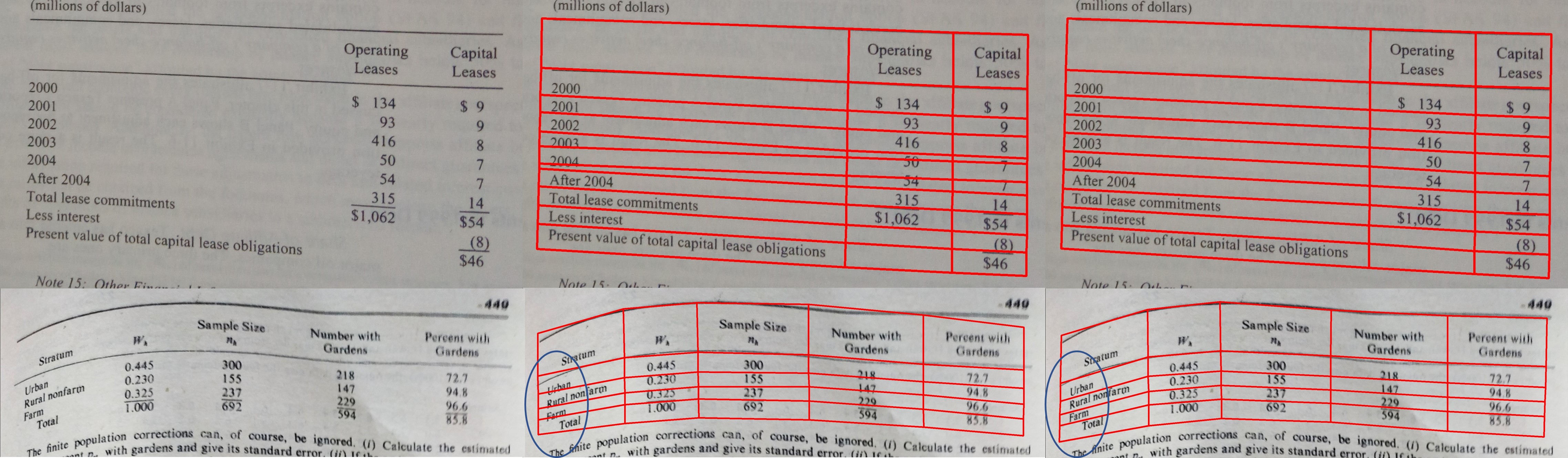}
    \caption{Qualitative results of SepRETR (middle) and DQ-DETR (right) on distorted tables.}
    \label{fig-7}
\end{figure}

\textbf{Effectiveness of DQ-DETR based split module.} As shown in Fig. \ref{fig-comp} and Fig. \ref{fig-7}, TSRFormer w/ DQ-DETR can achieve much higher localization accuracy on challenging tables. However, the gap between TSRFormer w/ DQ-DETR and TSRFormer w/ SepRETR in Table~\ref{tab:InHouse-05} is only 0.5\% in F1-score on our In-house dataset under the default setting. The reason is that the adopted evaluation metrics do not require high localization accuracy. Since the cTDaR TrackB metric uses the convex hull of all the text-line boxes located in each table cell to represent this cell, if no less than 50\% of one text-line located in a cell box, we assign the text-line to this cell for calculating convex hulls. The limitation of 50\% is not strict with localization accuracy since the predicted separation lines which are overlapped with some text-lines may not be punished. Therefore, increasing this number of limitation can generate a more strict evaluation metric. For instance, we replace 50\% with 80\% which means the predicted results will miss a text-line if less than 80\% of it is located in the related cell. The new experimental results are shown in Table~\ref{tab:Inhouse}. Under the new evaluation setting, the result of SPLERGE significantly drops by 10.2\% absolutely in F1-score. RobusTabNet and TSRFormer w/ SepRETR also drop 1.4\% and 1.1\% respectively. It's notable that the result of TSRFormer w/ DQ-DETR only drops 0.6\%, which is 1.0\% better than that of TSRFormer w/ SepRETR. Table~\ref{tab:sep-dq} shows that the DQ-DETR based split module does not exhibit a competitive advantage over the datasets with only horizontal-vertical tables such as SciTSR, PubTabNet, and FinTabNet. However, the DQ-DETR based split module has demonstrated its superiority over the SepRETR based split module on challenging table datasets such as WTW and the In-house dataset. Especially, TSRFormer w/ DQ-DETR achieves 2.3\% higher F1-score than TSRFormer w/ SepRETR on cTDaR TrackB2-Modern. We think the reason is that the training data is insufficient and DQ-DETR has better generalization ability.

\textbf{Ablation studies of several modules in TSRFormer.} As shown in Table \ref{tab:ablation1}, we analyze the influence of several modules in TSRFormer w/ both SepRETR and DQ-DETR and draw the following conclusions. First, SCNN \cite{spatialcnn} is important in both settings, which can bring significant gains. Second, due to the existence of spanning cells, adding a cell merging module and jointly training the split module with it for another 20 epochs can obtain consistent accuracy improvements. Third, following the conference paper \cite{lin2022tsrformer}, we add an additional auxiliary segmentation branch while training and find this can improve the SepRETR based split only model by 1.6\% in F1-score. However, after adding the cell merging module and training the models for another 20 epochs, the gap almost disappears. We find the reason is that this auxiliary segmentation branch can only accelerate the convergence of SepRETR based split module. In contrast, DQ-DETR is not significantly affected by this auxiliary branch because the progressive regression strategy in DQ-DETR significantly reduces the difficulty of this regression problem.

\begin{table}[!h]
\small
\setlength{\tabcolsep}{3mm} 
\setlength{\belowcaptionskip}{-0.1cm}
\caption{Ablation studies of several modules in TSRFormer on In-house dataset.}
\label{tab:ablation1}
\begin{center}
\begin{tabular}{cccccc}
    \hline\noalign{\smallskip}
      & SCNN & Aux-seg. & Cell Merging & F1. (\%)\\
    \noalign{\smallskip}
    \hline
    \multirow{4}{*}{\shortstack{SepRETR}}& &  & & 88.6\\
    & \checkmark & & & 91.0 \\
    & \checkmark & \checkmark  & &  92.6 \\
    & \checkmark & & \checkmark &  95.0\\
    & \checkmark & \checkmark & \checkmark & 95.2 \\
    \hline
    \multirow{4}{*}{\shortstack{DQ-DETR}}& &  & & 88.8 \\
    & \checkmark &  & & 92.5 \\
    & \checkmark & \checkmark & & 92.7 \\
    & \checkmark & & \checkmark & \textbf{95.7} \\
    & \checkmark & \checkmark & \checkmark & \textbf{95.7} \\
    \hline
\end{tabular}
\end{center}
\end{table}

\begin{table}[!h]
\centering
\caption{Effectiveness of progressive regression strategy.}
\label{tab:ablation-progressive}
\begin{tabular}{cc|l|l|l}
\hline
\multicolumn{2}{c|}{Method}                                & \multicolumn{1}{c|}{\multirow{2}{*}{Prec. (\%)}} & \multicolumn{1}{c|}{\multirow{2}{*}{Rec. (\%)}} & \multicolumn{1}{c}{\multirow{2}{*}{F1. (\%)}} \\ \cline{1-2}
\multicolumn{1}{c|}{SepRETR layers} & IterativeRETR layers & \multicolumn{1}{c|}{}                            & \multicolumn{1}{c|}{}                           & \multicolumn{1}{c}{}                         \\ \hline
\multicolumn{1}{c|}{0}              & 5                    & 94.4                                             & 94.0                                            & 94.2                                         \\
\multicolumn{1}{c|}{1}              & 4                    & 95.2                                             & 95.4                                            & 95.3                                         \\
\multicolumn{1}{c|}{2}              & 3                    & 95.1                                             & 95.3                                            & 95.2                                         \\
\multicolumn{1}{c|}{3}              & 2                    & 95.2                                             & 95.3                                            & 95.2                                         \\
\multicolumn{1}{c|}{4}              & 1                    & 95.0                                             & 95.0                                            & 95.0                                         \\
\multicolumn{1}{c|}{5}              & 0                    & 95.2                                             & 95.4                                            & 95.3                                         \\ \hline
\multicolumn{2}{c|}{5 DQ-DETR layers}                      & \textbf{95.5}                                             & \textbf{95.9 }                                           & \textbf{95.7 }                                        \\ \hline
\end{tabular}
\end{table}

\textbf{Effectiveness of progressive regression strategy}. 
We conduct several experiments to demonstrate the effectiveness of our proposed progressive regression strategy in DQ-DETR. To do so, we compare its performance with the direct regression strategy (i.e., SepRETR) presented in our conference paper \cite{lin2022tsrformer} and the iterative refinement strategy adopted by DINO \cite{zhang2022dino}. The decoder layer for iterative refinement, named \textbf{IterativeRETR} in our work, maintains the same architecture as the DQ-DETR decoder layer, except for the progressive generation of new points. This decoder layer also enhances the embedding of queries output by the preceding decoder layer, followed by refining the positions of the corresponding points. To comprehensively compare our proposed progressive regression strategy with the other two strategies, we stack several SepRETR layers with several IterativeRETR layers to replace the DQ-DETR decoder for separation line prediction. 
Table~\ref{tab:ablation-progressive} illustrates that our proposed DQ-DETR decoder achieves better performance than various combinations of SepRETR and IterativeRETR layers when using the same number of decoder layers. This result highlights the superiority of our progressive regression strategy over the direct regression and iterative refinement strategies.

\textbf{Effectiveness of prior-enhanced bipartite matching strategy.} We conduct several experiments by training the DQ-DETR based split module with different matching strategies and epochs. As shown in Table \ref{tab:ablation3}, training the model with the original strategy in DETR by 40 epochs achieves much higher accuracy than training by 20 epochs, which means the split module has not fully converged. In contrast, using the proposed prior-enhanced matching strategy can achieve much better results. The small performance gap between models trained with 20 and 40 epochs shows that these two models have converged well, which demonstrates that our prior-enhanced matching strategy can make convergence much faster.

\begin{table}[ht]
\small
\setlength{\tabcolsep}{2mm} 
\setlength{\belowcaptionskip}{-0.1cm}
\caption{Effectiveness of prior-enhanced matching strategy.}
\label{tab:ablation3}
\begin{center}
\begin{tabular}{ccc}
    \hline\noalign{\smallskip}
    Matching Strategy & \#Epochs & F1. (\%)\\
    \noalign{\smallskip}
    \hline
    \noalign{\smallskip}
    Original in DETR & 20 & 88.1 \\
    Prior-enhanced & 20 & \textbf{92.7} \\
    Original in DETR & 40 &  90.2 \\
    Prior-enhanced & 40 & \textbf{92.8} \\
    \hline
\end{tabular}
\end{center}
\end{table}

\begin{figure}[htbp]
    \centering
    \subfigure[Failure case of the excessively curved tables.]{
        \includegraphics[height=7cm]{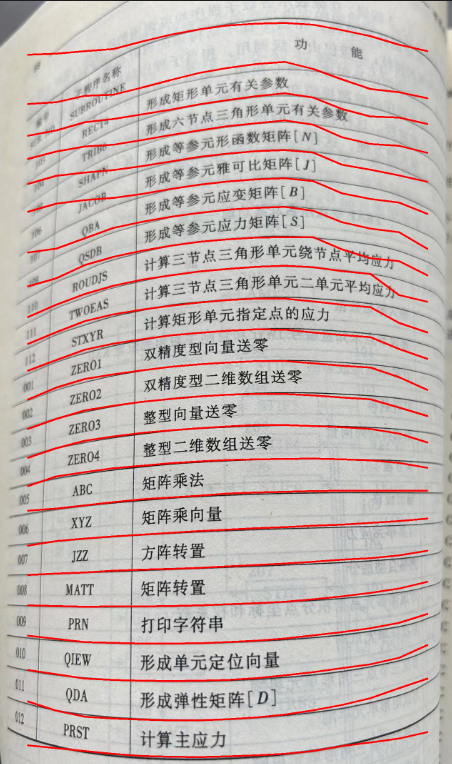}
    }
    \subfigure[Failure case of the extremely dense tables.]{
        \includegraphics[width=7cm]{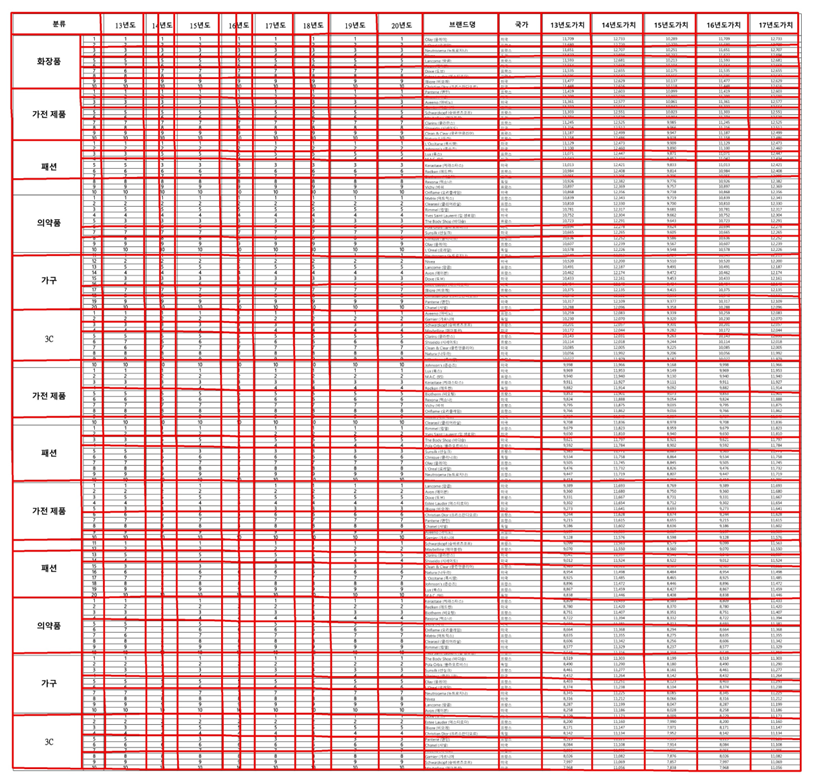}
    }
    \centering
    \caption{Some typical failure cases, including extremely curved table and extremely dense table.}
    \label{fig:failure_cases}
\end{figure}

\subsection{Analysis on the efficiency of our approach}
We make a thorough analysis on the efficiency of our proposed approach over our in-house dataset, including model parameters, GFOLPS, FPS, and GPU memory footprint, as shown in Table \ref{tab:efficiency}. Compared with the competitive method RobusTabNet \cite{ma2022robust}, although the configuration of our proposed TSRFormer with DQ-DETR has about 75\% more parameters and 20\% less FPS than RobusTabNet due to the usage of multiple deformable transformer decoder layers, it achieves 4.2\% higher F1-score than the former one on In-house dataset. We also design the lightweight version of SepRETR \cite{lin2022tsrformer} and DQ-DETR to explore the relationship between efficiency and performance. Specifically, for light-SepRETR, we reduce the point number K predicted for each curvilinear line from 15 to 9 and change the query dimension, head number, and dimension of feed-forward layers in the transformer decoder from 256, 16, 1024 to 128, 8, 512, respectively. For light-DQ-DETR, we keep the same configuration as light-SepRETR with the exception of the point number K, which is reduced from 15 to 11. As shown in Table \ref{tab:efficiency}, the parameters, GFLOPS, and FPS of light-DQ-DETR based TSRFormer are very close to that of RobusTabNet yet light-DQ-DETR based TSRFormer still achieves 3.6\% higher F1-score.
Compared with light-SepRETR based TSRFormer, light-DQ-DETR improves the F1-score by 1.5\%, but the cost of these two models is similar.

\begin{table}[!h]
\small
\setlength{\tabcolsep}{2mm} 
\setlength{\belowcaptionskip}{-0.1cm}
\caption{Analysis on the efficiency of our proposed approach on In-house dataset with Limitation threshold = 80\%. }
\label{tab:efficiency}
\begin{center}
\begin{tabular}{l|l|l|l|l|l}
\hline Model & \#Param(M) & GFLOPS & F1. (\%) & FPS & $\begin{array}{l}\text { GPU Memory } \\
\text { Footprint (G) }\end{array}$ \\
\hline RobusTabNet & 20.1 & 26.42 & $90.9$ & 5.19 & 2.3 \\
\hline TSRFormer(SepRETR) & 26.7 & 36.86 & $94.1$ & 5.17 & 5.5 \\
\hline TSRFormer(light-SepRETR) & 21.7 & 30.35 & $93.0$ & 6.71 & 3.1 \\
\hline TSRFormer(DQ-DETR) & 35.0 & 35.15 & $95.1$ & 4.17 & 5.8 \\
\hline TSRFormer(light-DQ-DETR) & 22.9 & 28.01 & $94.5$ & 5.34 & 3.4 \\
\hline
\end{tabular}
\end{center}
\end{table}

\subsection{Limitations of our approach}
Although the proposed TSRFormer with DQ-DETR demonstrates superior capability in most scenarios as demonstrated in the previous experiments, it still has some limitations. For example, the progressive regression approach still struggles with excessively curved tables due to the potential cumulative error problem and the lack of training data. Furthermore, the regression-based TSR approach still fails on some extremely dense tables with extreme size. An adaptive scaling strategy may be necessary for tables with extreme sizes. Some failure examples are presented in Fig.~\ref{fig:failure_cases}. Note that these difficulties are common challenges faced by other state-of-the-art methods. Finding practical solutions to these problems will be the focus of our future work.

\section{Conclusion and future work}
In this paper, we presented TSRFormer, a new regression based separation line prediction approach for table structure recognition, which contains two effective components: a DQ-DETR based split module for separation line prediction and a relation network based cell merging module for spanning cell recovery. Compared with previous image segmentation based separation line detection methods, our DQ-DETR based separation line regression approach can achieve higher TSR accuracy without relying on heuristic mask-to-line modules. Experimental results show that the proposed prior-enhanced bipartite matching strategy can accelerate the convergence speed of two-stage DETR effectively. Furthermore, dynamic query and progressive regression approaches are beneficial for high localization accuracy compared to SepRETR. Consequently, our approach has achieved state-of-the-art performance on four public benchmarks, including SciTSR, PubTabNet, FinTabNet, WTW, and cTDaR TrackB2-Modern. We have further validated the robustness of our approach to tables with complex structures, borderless cells, large blank spaces, empty or spanning cells as well as distorted or curved shapes on a more challenging real-world In-house dataset.

For future work, we will study how to improve the progressive regression approach to reduce the potential cumulative error problem. Furthermore, to achieve more robust structure recognition of dense tables, we will study effective technologies for adaptive scaling. The concept of Dynamic Query is general and elegant for regression problems. In the future, we will explore its full potential on other regression tasks.


 \bibliographystyle{elsarticle-num} 
 \biboptions{numbers,sort&compress}
 \bibliography{bibfile}





\end{document}